\documentclass{article} 
\usepackage{iclr2027_conference,times}

\usepackage{amsmath,amsfonts,bm}

\def\eqref#1{equation~\ref{#1}}

\def\1{\bm{1}}

\DeclareMathAlphabet{\mathsfit}{\encodingdefault}{\sfdefault}{m}{sl}
\SetMathAlphabet{\mathsfit}{bold}{\encodingdefault}{\sfdefault}{bx}{n}

\usepackage[hidelinks]{hyperref}
\usepackage{url}

\usepackage{graphicx}
\usepackage{amsmath} 
\usepackage{multirow}
\usepackage{algorithm}
\usepackage{algpseudocode}
\usepackage{caption}
\usepackage{booktabs}
\usepackage[dvipsnames]{xcolor}
\usepackage{siunitx}
\usepackage{pifont}
\newcommand{\cmark}{\ding{51}}
\newcommand{\xmark}{\ding{55}}
\usepackage{amssymb}
\usepackage{wrapfig}

\title{ViCoR: Reliable Molecular Structure Extraction via Spatially Aligned Verification and Executable Revision}

\author{
\textbf{Yujian Yuan}$^{1}$, \quad
\textbf{Xin Cai}$^{2}$, \quad
\textbf{Yufan Chen}$^{1}$, \quad
\textbf{Jiaxin Xu}$^{1}$, \quad
\textbf{Mengdi Liu}$^{3}$, \quad
\textbf{Zhichao Tan}$^{1}$,
\\[0.15em]
\textbf{Long Chen}$^{1}$, \quad
\textbf{Hanyu Gao}$^{1,*}$
\\
$^{1}$The Hong Kong University of Science and Technology
\quad
$^{2}$The Chinese University of Hong Kong
\\
$^{3}$ Institute of Computing Technology, Chinese Academy of Sciences.
\\
\texttt{yyuanbn@connect.ust.hk, hanyugao@ust.hk, *:Corresponding author}
}

\iclrfinalcopy

\iclrfinalcopy 
\begin{document}

\maketitle

\begin{abstract}

Reliable optical chemical structure recognition (OCSR) is essential for
building high-quality chemical data from scientific literature, yet even small
recognition errors can propagate into chemical databases and downstream
models.
In practice, recognized structures often require manual inspection and
correction before use, making large-scale data curation costly and difficult to
scale.
We therefore study \emph{Selective Structure Recognition (SSR)}, a
post-recognition setting that automatically produces reliable structured
outputs while rejecting unresolved cases.
Selection-only approaches can improve reliability by rejection, but cannot
create additional correct outputs beyond those produced by the base recognizer.
We propose \textbf{ViCoR}, a \emph{repair-before-rejection} framework for
iterative \textbf{V}er\textbf{I}fi\textbf{C}ati\textbf{O}n and \textbf{R}evision.
Its key idea is to make observation--prediction correspondence explicit:
coordinate-preserving rendering establishes spatial correspondence between the
source image and predicted structure, while index anchoring maps localized
visual discrepancies to executable graph edits without full-structure
regeneration.
A shared VLM is progressively trained from verification to revision.
On two real-world OCSR benchmarks, ViCoR improves overall accuracy from
73.53\% to 88.26\% and from 61.83\% to 84.32\%, while achieving over 97\%
accepted accuracy at 85--89\% coverage.
The resulting molecular data further improve reaction-extraction F1 by 15.5 and literature-sourced reaction prediction accuracy by 7.7 and 5.8 points, demonstrating
the value of automated reliability control for scientific data curation and
downstream chemical learning. Codes to be released.

\end{abstract}

\section{Introduction}

Extracting molecular structures from chemical literature and patents is fundamental to building machine-readable chemical knowledge bases~\citep{fang2025molparser,fan2024openchemie}. Because molecular structures are predominantly communicated as two-dimensional depictions, optical chemical structure recognition (OCSR) is used to convert these images into structured representations such as SMILES~\citep{weininger1988smiles} or molecular graphs~\citep{qian2023molscribe}. Recent OCSR systems have achieved strong recognition performance, enabling increasingly large-scale extraction from scientific documents~\citep{morin2024patcid}.
However, recognition errors remain inevitable in real-world deployment at scale, and even a single incorrect atom or bond can change the identity of the entire
molecule.
Such errors often require costly and time-consuming human correction before extracted structures can be reliably used in chemical databases or downstream model training~\citep{tan2026data, nippa2024enabling}.
This raises a practical question beyond improving the recognizer itself: \textit{can we automatically turn imperfect OCSR predictions into reliable molecular data?}

\begin{figure}[t]
    \centering
    \includegraphics[width=1.0\textwidth]{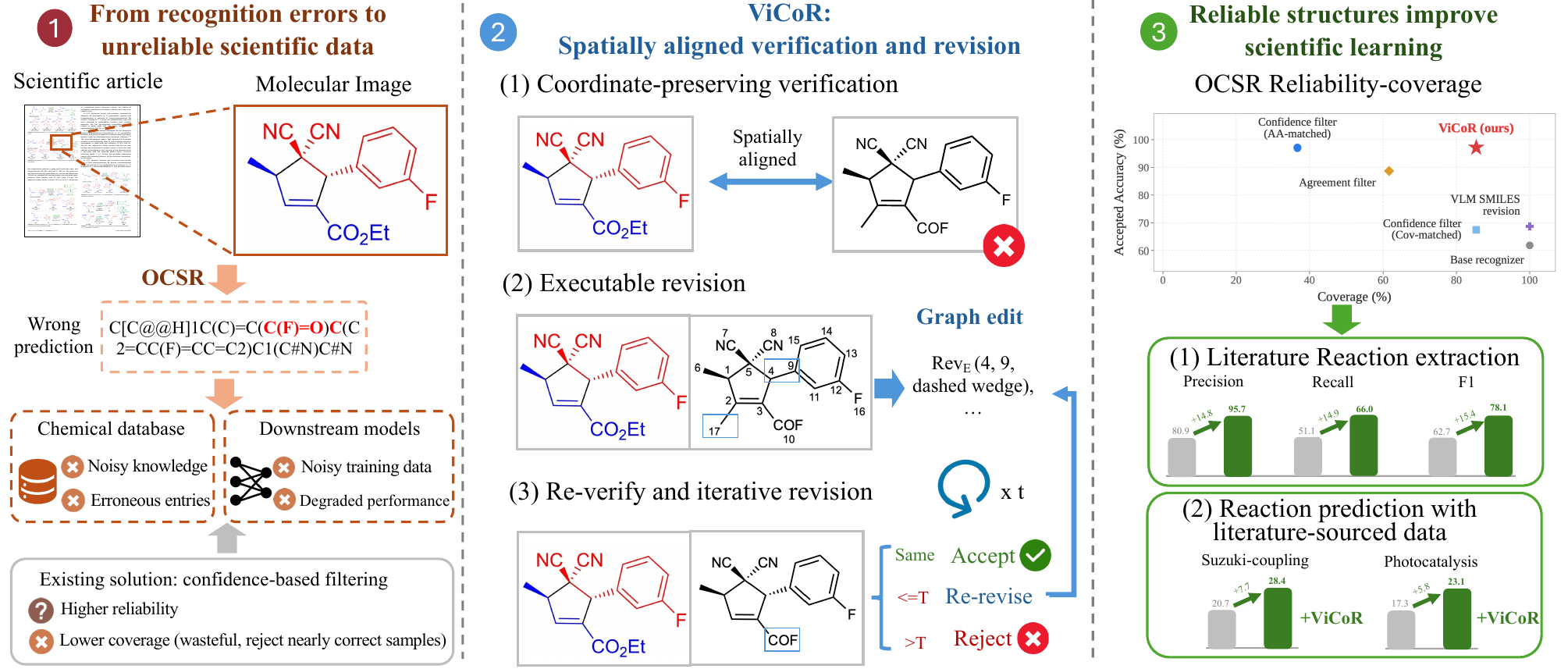}
    \vspace{-2 em}
    \caption{
    \textbf{From unreliable OCSR outputs to reliable scientific data with ViCoR.}  Recognition errors can corrupt chemical data.
    ViCoR uses spatially aligned verification and executable revision to recover correct structures, improving both reliability–coverage and downstream chemical learning.}
    \label{fig:fig1}
    \vspace{-1 em}
\end{figure}

A natural post-recognition solution is to identify unreliable predictions and reject them before downstream use.
However, because selection can only retain or discard predictions already
produced by recognizer, improving reliability comes at the
cost of reduced coverage.
This limitation is particularly wasteful for OCSR, where many recognition errors are localized to only a few atoms or bonds while most of the structure is already correct.
We therefore study \emph{Selective Structure Recognition (SSR)}, a
post-recognition setting that seeks reliable structured outputs while allowing unresolved cases to be rejected.
Rather than relying on rejection alone, we pursue 
\emph{repair-before-rejection} principle: repair incorrect predictions when
possible, and reject them only when they remain unresolved.

The key challenge is not merely determining whether a structured prediction is incorrect, but grounding the error in the structure.
This requires comparing the prediction with the source image, localizing the visual inconsistency, and associating it with the corresponding structured
entity.
However, the two representations are not naturally aligned:
SMILES does not preserve the spatial layout of the molecule, while a molecular
graph does not explicitly associate its atoms and bonds with visual regions.
Reliable verification and localized correction therefore share a central
challenge: establishing explicit correspondence between the source observation
and the structured prediction.

To address these challenges, we introduce \textbf{ViCoR}, a post-recognition
framework that makes observation--prediction correspondence explicit.
ViCoR re-renders the current prediction using its recognized spatial layout,
turning cross-representation alignment into direct spatial comparison with
the source image.
For revision, ViCoR augments the rendered prediction with atom indices,
providing explicit anchors that associate localized visual discrepancies with
specific graph entities and executable graph edits.
Together, these designs establish a grounded correction path from
\emph{visual inconsistency} to \emph{graph entity} to \emph{structural edit},
allowing incorrect regions to be revised while preserving the remaining
structure.
The revised prediction is then iteratively re-rendered and re-verified, enabling multiple
localized errors to be resolved before unresolved cases are rejected.
Verification and revision share this aligned visual interface and are
implemented by a single VLM.

We evaluate ViCoR on two newly curated in-the-wild benchmarks and eight
standard OCSR benchmarks.
On real-world benchmarks, ViCoR raises overall accuracy from
73.53\% to 88.26\% and from 61.83\% to 84.32\%, while retaining over
97\% accepted accuracy at 85--89\% coverage.
These gains translate to downstream chemical tasks, improving reaction-extraction F1 by 15.45 and literature-sourced reaction-prediction exact match by 7.7/5.8 points on
Suzuki coupling/photocatalysis.

In summary, our main contributions are highlighted as follows:
\vspace{-1 em}

\begin{itemize}

    \item We formulate \emph{Selective Structure Recognition (SSR)} as a
    post-recognition problem for reliable structure prediction, and show that
    selection-only methods cannot produce more correct data than base
    recognizer already predicts, motivating \emph{repair-before-rejection}
    strategy.

\item We introduce \textbf{ViCoR}, which repurposes spatial information from
the initial structured prediction to establish explicit correspondence with
the source image.
Coordinate-preserving rendering enables direct spatial comparison, while index
anchoring maps localized discrepancies to executable graph edits, forming an iterative verification--revision process.

    \item Extensive experiments across ten OCSR benchmarks show that ViCoR
    improves both recognition accuracy and the reliability--coverage trade-off.
    The recovered molecular structures further improve reaction extraction and
    reaction prediction from literature-sourced data, demonstrating their value
    for constructing reliable scientific datasets.
    
\end{itemize}

\section{Related Work}
\vspace{-1em}

\begin{figure*}[t]
    \centering
    \includegraphics[width=1.0\textwidth]{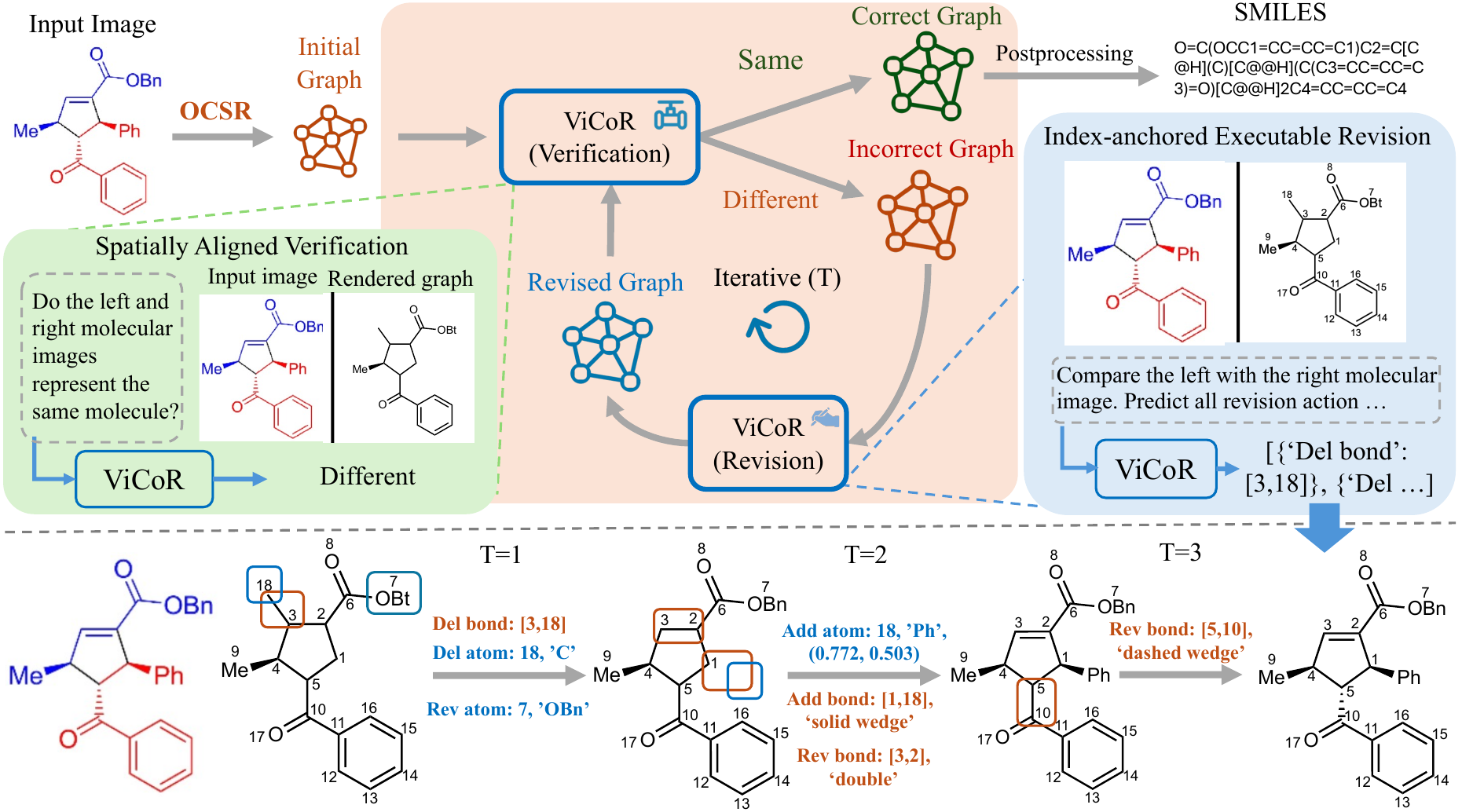}
    \vspace{ -2 em}
    \caption{\textbf{Overview of ViCoR.} Starting from an OCSR-predicted initial graph, ViCoR iteratively verifies and revises the prediction against the source image. Coordinate-preserving rendering establishes spatial correspondence for verification, while atom indexing maps localized visual discrepancies to executable graph edits for revision. Each revised graph is re-rendered and re-verified until it is accepted or reaches the maximum iteration T. 
    }
    \label{fig:overview}
    \vspace{ -0.5 em}
\end{figure*}

\textbf{Optical Chemical Structure Recognition (OCSR).}
OCSR converts molecular images into
structured representations.
Existing methods range from rule-based pipelines~\citep{peryea2019molvec,filippov2009optical}
to end-to-end SMILES generation~\citep{fang2025molparser,rajan2020decimer} and
graph-based recognition~\citep{qian2023molscribe,morin2023molgrapher,chen2024molnextr}.
VLMs have also been explored for broader chemical figure understanding and
reaction extraction~\citep{chen2025towards,song2025rxncaption,chen2025multi,leong2025mermaid}.
Despite strong recognition performance, the remaining structural errors often
require manual inspection and correction before extracted molecules can be
reliably used in chemical databases or downstream learning.
Our work targets this complementary post-recognition stage, aiming to
automatically produce reliable structured outputs from imperfect OCSR
predictions while rejecting unresolved cases.

\textbf{Selective Prediction and Post-Recognition Refinement.}
Selective prediction improves reliability by rejecting uncertain outputs,
leading to the classical risk--coverage trade-off
~\citep{el2010foundations,geifman2017selective,geifman2019selectivenet}.
Recent multimodal methods reduce unnecessary abstention by gathering additional
visual evidence, but still leave the underlying structured prediction unchanged
~\citep{srinivasan2024selective}.
A complementary line of work refines model outputs through critique,
verification, or iterative feedback
~\citep{madaan2023self,gou2024critic},
primarily for free-form generation.
Image-to-structure correction poses an additional challenge: a localized visual
discrepancy must be associated with the corresponding structured entity and
translated into a valid edit.
ViCoR makes this correspondence explicit through coordinate-preserving
alignment and index-anchored executable graph edits, while rejecting cases that
remain unresolved.

\section{Method}
\vspace{ -0.5 em}

We first formulate \emph{Selective Structure Recognition (SSR)} as a
post-recognition problem that seeks reliable structured outputs from the
predictions of an existing recognizer.
To move beyond selection-only rejection, we propose \textbf{ViCoR}, which
follows a \emph{repair-before-rejection} principle and iteratively verifies and
revises incorrect predictions before rejecting unresolved cases.
For OCSR, ViCoR establishes explicit spatial correspondence through
coordinate-preserving rendering and uses index anchoring to map localized
visual discrepancies to executable graph edits.
Finally, we progressively train a shared VLM from verification to revision to
support both operations in the same framework.

\subsection{Task Definition: Selective Structure Recognition}
\label{sec:problem_setup}

We study \emph{Selective Structure Recognition (SSR)}, a post-recognition setting that operates on structured predictions produced by an existing recognizer.
Given an initial prediction together with its source observation, SSR seeks to produce a reliable structured output for downstream use, while rejecting cases that cannot be resolved reliably.

In this work, we instantiate SSR for optical chemical structure recognition (OCSR)~\citep{qian2023molscribe}.
Given a set of molecular images $\{I_k\}_{k=1}^{N}$, an base OCSR recognizer $\mathcal{A}$ first the predicts initial molecular structure $Y_k^{(0)}\in\mathcal{Y}$ for each image:
\begin{equation}
    Y_k^{(0)}=\mathcal{A}(I_k),
\end{equation}
A post-recognition system $\mathcal{F}$ then processes each
image--prediction pair and returns a final candidate together with an
acceptance decision:
\begin{equation}
    (\widetilde{Y}_k,a_k)
    =
    \mathcal{F}(I_k,Y_k^{(0)}),
    \qquad
    \widetilde{Y}_k\in\mathcal{Y},
    \quad
    a_k\in\{0,1\},
\end{equation}
where $a_k=1$ denotes acceptance and $a_k=0$ denotes rejection.
Let $Y_k^\star\in\mathcal{Y}$ denote the ground-truth structure.
We evaluate $\mathcal{F}$ using three complementary metrics:
\begin{equation}
\label{eq:ssr_metrics}
\mathrm{OA}
=
\frac{1}{N}\sum_{k=1}^{N}\mathbb{I}[\widetilde{Y}_k=Y_k^\star],
\quad
\mathrm{AA}
=
\frac{\sum_{k=1}^{N}a_k\mathbb{I}[\widetilde{Y}_k=Y_k^\star]}
{\sum_{k=1}^{N}a_k},
\quad
\mathrm{Cov}
=
\frac{1}{N}\sum_{k=1}^{N}a_k .
\end{equation}
Overall Accuracy (OA) measures the accuracy of the final structured outputs
over the full input set, while Accepted Accuracy (AA) measures the reliability
of accepted predictions and Coverage (Cov) measures the fraction of inputs
accepted.
Since the preferred reliability--coverage trade-off depends on downstream
requirements, SSR does not assume a universally optimal operating point.
We say that $\mathcal{F}_a$ \emph{Pareto-dominates} $\mathcal{F}_b$,
denoted by $\mathcal{F}_a \succ \mathcal{F}_b$, if
\begin{equation}
\label{eq:dominance}
    \mathrm{AA}(\mathcal{F}_a)
    \ge
    \mathrm{AA}(\mathcal{F}_b),
    \qquad
    \mathrm{Cov}(\mathcal{F}_a)
    \ge
    \mathrm{Cov}(\mathcal{F}_b),
\end{equation}
with at least one inequality being strict.
SSR therefore seeks a better reliability--coverage trade-off, balancing high reliability with broad coverage.

\subsection{Iterative Verification and Revision}
\label{sec:iterative_framework}

A natural approach to SSR is \emph{selection-only post-recognition}: accept the judged-reliable predictions and reject the rest without changing their
structured outputs.

\textbf{Selection-only ceiling.}
Consider a selection-only system $\mathcal{F}_{\mathrm{sel}}$, for which
$\widetilde{Y}_k=Y_k^{(0)}$ for every sample.
Since selection cannot turn an incorrect prediction into a correct one,
\begin{equation}
\label{eq:selection_ceiling}
\begin{aligned}
\mathrm{AA}\!\cdot\!\mathrm{Cov}
&=
\frac{1}{N}\sum_{k=1}^{N}
a_k\,\mathbb{I}\!\left[Y_k^{(0)}=Y_k^\star\right]
\leq \mathrm{OA}^{(0)}, \\
\mathrm{AA}\geq\tau
&\Longrightarrow
\mathrm{Cov}
\leq
\min\!\left(1,\frac{\mathrm{OA}^{(0)}}{\tau}\right).
\end{aligned}
\end{equation}
Thus, with the base accuracy $\mathrm{OA}^{(0)}$ fixed, the maximum attainable
coverage at any target reliability $\tau$ is also bounded.
Selection can only move along this reliability--coverage trade-off.
Improving the attainable coverage at a given reliability therefore requires
increasing the overall number of correct predictions.
This motivates repairing incorrect predictions before rejecting unresolved
cases.

\textbf{Repair Before Rejection.}
OCSR provides an opportunity to move beyond the selection-only ceiling.
Many recognition errors are localized to a small number of atoms or bonds,
while most of the predicted structure remains correct~\citep{chen2024molnextr}.
Rather than directly rejecting such predictions, ViCoR follows a
\emph{repair-before-rejection} principle: it first attempts to revise
correctable errors and rejects a prediction only if it remains unresolved.
ViCoR realizes this principle through verification and revision.
Given an input image $I$ and the current prediction $Y^{(t)}$, the verifier
$\mathcal{V}$ determines whether the prediction is consistent with the source
observation:
\begin{equation}
    \mathcal{V}(I,Y^{(t)})\in\{0,1\}.
\end{equation}
If verified, $Y^{(t)}$ is accepted.
Otherwise, the reviser $\mathcal{R}$ predicts a structured update
\begin{equation}
\label{eq:revision}
    \Delta^{(t)}
    =
    \mathcal{R}(I,Y^{(t)}),
    \qquad
    Y^{(t+1)}
    =
    Y^{(t)}\oplus\Delta^{(t)},
\end{equation}
where $\oplus$ applies the predicted update to the current structure.

\begin{wrapfigure}{r}{0.5\columnwidth}
  \vspace{-0.8em}
  \begin{minipage}{\linewidth}
    \footnotesize

    \rule{\linewidth}{0.4pt}
    \vspace{-0.35em}

    \captionsetup{
      type=algorithm,
      justification=raggedright,
      singlelinecheck=false,
      skip=2pt
    }
    \caption{Iterative verification and revision.}
    \label{alg:iterative}

    \vspace{-0.45em}
    \rule{\linewidth}{0.4pt}
    \vspace{0.15em}

    \begin{algorithmic}[1]
      \State \textbf{Input:} $I,Y^{(0)},T$
      \For{$t=0,\ldots,T$}
        \If{$\mathcal{V}(I,Y^{(t)})=1$}
          \State \Return $(Y^{(t)},1)$
        \EndIf
        \If{$t<T$}
          \State $Y^{(t+1)} \leftarrow
          Y^{(t)} \oplus \mathcal{R}(I,Y^{(t)})$
        \EndIf
      \EndFor
      \State \Return $(Y^{(T)},0)$
    \end{algorithmic}

    \vspace{0.15em}
    \rule{\linewidth}{0.4pt}
  \end{minipage}
  \vspace{-0.7em}
\end{wrapfigure}

\textbf{Iterative Verification and Revision.}
Since challenging samples often contain multiple localized errors, a single-trial revision is usually insufficient to perfectly repair the entire structure.
ViCoR therefore re-verifies each revised prediction and performs another revision only if inconsistencies remain.
As shown in Alg.~\ref{alg:iterative}, verification and revision alternate from
$Y^{(0)}$ until the prediction is verified or the maximum of $T$ revisions is
reached.
Verification therefore serves both as the acceptance gate and an adaptive
stopping criterion: verified predictions are accepted immediately, while
unresolved ones proceed to further revision.
Predictions that remain unverified after $T$ revisions are rejected.

\subsection{Spatially Aligned Verification and Index-Anchored Revision}
\label{sec:spatial_revision}

Verification and revision require comparing a structured prediction with its
source image, yet the two are represented in different spaces.
Typical OCSR outputs such as SMILES discard spatial layout, while molecular
graphs encode structure without explicitly associating atoms and bonds with
visual regions.
ViCoR closes this representation gap in image space: coordinate-preserving
rendering establishes spatial correspondence for verification, while atom
indexing makes the aligned graph entities directly addressable for revision.

\textbf{Spatially Aligned Visual Verification.}
At each iteration, we represent the current structured prediction $Y^{(t)}$ as a
molecular graph
\begin{equation}
    G=(V,E,X),
\end{equation}
where $V=\{v_i\}_{i=1}^{n}$ denotes the atoms,
$E\subseteq V\times V$ denotes the bonds, and
$X=\{(x_i,y_i)\}_{i=1}^{n}$ contains the corresponding two-dimensional atom locations
provided by the graph-based OCSR recognizer.
Each atom $v_i$ has a label $\ell_i\in\mathcal{L}$, and each edge
$(v_i,v_j)\in E$ has a bond type $b_{ij}\in\mathcal{B}$.

ViCoR renders the predicted graph using its recognized coordinates:
\begin{equation}
    \widetilde{I}
    =
    \rho(G;\mathbf{X}),
\end{equation}
where $\rho$ denotes coordinate-preserving molecular rendering.
Unlike canonical-layout rendering, which may rearrange the molecular
structure, coordinate-preserving rendering keeps corresponding atoms and bonds
at approximately matching locations in the source and rendered images.

We then place the source image and rendered prediction side by side and use the
resulting interface for verification:
\begin{equation}
    P(I,G)=I\,\Vert\,\widetilde{I},
    \qquad
    \mathcal{V}\!\left(P(I,G)\right)\in\{0,1\},
\end{equation}
where $\Vert$ denotes side-by-side presentation.
This keeps both views separately visible while placing spatially corresponding
regions in a shared visual context, enabling direct spatial comparison between
the source image and the predicted structure.

\textbf{Index-Anchored Executable Revision.}
Verification only needs to determine whether an inconsistency exists, whereas
revision must additionally identify the graph entity associated with it.
We therefore overlay a unique index on each predicted atom and construct
\begin{equation}
    \widetilde{I}_{\mathrm{idx}}
    =
    \rho_{\mathrm{idx}}(G;\mathbf{X}),
    \qquad
    P_{\mathrm{idx}}(I,G)
    =
    I \,\Vert\, \widetilde{I}_{\mathrm{idx}}.
\end{equation}
The atom indices make graph entities explicitly addressable, allowing a
localized visual discrepancy to be mapped to a specific atom or bond.
Rather than regenerating the complete molecular structure, the reviser takes
the indexed interface as input and predicts a sequence of executable graph
edits:
\begin{equation}
\label{eq:revision_output}
    \mathcal{R}(I,G)
    =
    \mathcal{R}\!\left(P_{\mathrm{idx}}(I,G)\right)
    =
    \Delta
    =
    (\delta_1,\ldots,\delta_K).
\end{equation}
We define six edit types covering node- and edge-level changes:
\begin{align}
    \Omega_{\mathrm{node}}
    &=
    \left\{
    \mathrm{Add}_V(i,\ell,(x,y)),
    \mathrm{Del}_V(i),
    \mathrm{Rev}_V(i,\ell')
    \right\},
    \label{eq:node_edits}\\
    \Omega_{\mathrm{edge}}
    &=
    \left\{
    \mathrm{Add}_E(i,j,b),
    \mathrm{Del}_E(i,j),
    \mathrm{Rev}_E(i,j,b')
    \right\}.
    \label{eq:edge_edits}
\end{align}
Here, $i$ and $j$ denote atom indices,
$\ell,\ell'\in\mathcal{L}$ denote atom labels,
$b,b'\in\mathcal{B}$ denote bond types, and
$(x,y)\in\mathbb{R}^{2}$ denotes the normalized location of an inserted atom.
The update operator $\oplus$ applies the predicted edits sequentially and
skips any edit invalidated by earlier graph modifications.
Further details are provided in the Appendix.

\subsection{Verification-to-Revision Progressive Training}
\label{sec:progressive_training}

Verification and revision are built on the same spatially aligned comparison
between the source image and the predicted structure.
The main difference is that verification only determines whether the two are
consistent, whereas revision further identifies the inconsistency and translates
it into a structural edit.
Based on the shared visual basis, we propose to train a single VLM $\mathcal{M}$,
with task-specific prompts controlling the two behaviors:
\begin{equation}
    \mathcal{V}(I,G)
    =
    \mathcal{M}\!\left(P(I,G),p_{\mathrm{ver}}\right),
    \qquad
    \mathcal{R}(I,G)
    =
    \mathcal{M}\!\left(P_{\mathrm{idx}}(I,G),p_{\mathrm{rev}}\right).
\end{equation}
More importantly, this asymmetry motivates our
\emph{verification-to-revision progressive training}: we first train the model
to recognize structural discrepancies through verification, and then transfer
this learned correspondence to the more fine-grained revision task.

\textbf{Stage 1: Verification-Oriented Pretraining.}
We first train $\mathcal{M}$ on the spatially aligned interface $P(I,G)$ to
distinguish structurally consistent predictions from inconsistent ones.
By varying drawing styles and image noise during training, the model learns to
ignore appearance differences unrelated to molecular structure and focus on
structure-relevant discrepancies.
Meanwhile, verification encourages robust spatial correspondence between the
paired images, providing a foundation for subsequent localized revision.

\textbf{Stage 2: Revision-Oriented Fine-Tuning.}
Starting from the verification-pretrained model, we further train $\mathcal{M}$
on the indexed interface $P_{\mathrm{idx}}(I,G)$ to generate executable graph
edits.
This extends the learned correspondence from deciding \emph{whether} an
inconsistency exists to identifying \emph{where} it occurs and \emph{how} the
corresponding graph entity should be revised.

\section{Experiment}

\textbf{Training Data Synthesis.}
We construct 1.3M synthetic training samples from PubChem molecules~\citep{kim2016pubchem}, including 800K for verification-oriented pretraining and 500K for revision-oriented fine-tuning.
For each sample, the source image is generated with Indigo using randomized drawing styles and image noise, while the candidate graph is rendered by RDKit~\citep{rdkit2016} under the same coordinate frame with stereochemistry preserved.
To better mimic realistic OCSR errors, we also include incorrect predictions produced by MolScribe on synthesized molecular images.
Further details are provided in the Appendix.

\textbf{Evaluation Datasets.}
We evaluate ViCoR on eight standard OCSR benchmarks, including six realistic
datasets (CLEF~\citep{piroi2010clef}, UOB~\citep{sadawi2012chemical},
JPO~\citep{fujiyoshi2011robust}, USPTO~\citep{filippov2009optical},
Staker~\citep{staker2019molecular}, and ACS~\citep{qian2023molscribe})
and two synthetic datasets (Indigo and ChemDraw~\citep{qian2023molscribe}).
To evaluate OCSR under recent journal-domain conditions, we curate two
in-the-wild benchmarks, OCIE and JIE.
They contain low-resolution journal crops with diverse drawing styles, color,
visual distractors, and naturally co-occurring challenges such as R-groups,
complex abbreviations, and disconnected structures.
OCIE contains 1,141 molecular images cropped from the OpenChemIE test set
~\citep{fan2024openchemie}, while JIE contains 1,779 molecules cropped from
243 recent journal figures (e.g., \emph{JACS}, \emph{Nature}).

\textbf{Metrics.}
We evaluate SSR using the three metrics defined in
Eq.~\ref{eq:ssr_metrics}: Overall Accuracy (OA), Accepted Accuracy (AA), and
Coverage (Cov).
OA measures exact-match accuracy over the full test set, while AA and Cov
measure the reliability and coverage of accepted predictions, respectively.

\textbf{Implementation Details.}
We use Qwen2.5-VL-3B~\citep{bai2025qwen2} as the shared VLM and fine-tune it
with LoRA~\citep{hu2022lora}.
The source and rendered images are horizontally concatenated as a single
side-by-side input. Normalized insertion coordinates are quantized into 1,000
location tokens, and predicted graph edits are executed sequentially by
$\oplus$.
We train with batch size 64 using AdamW with weight decay $0.1$ for 2 epochs
of stage 1 pretraining followed by 1 epoch of
stage 2 fine-tuning.
We use cosine learning-rate decay with a peak rate of $2\times10^{-5}$ and
100 warm-up steps.
All experiments are conducted on 8 NVIDIA RTX 3090 GPUs.

\begin{table*}[t]
  \footnotesize
  \centering
  \caption{Selective Structure Recognition on real-world OCSR benchmarks.
  OA/AA denote exact-match accuracy over all/accepted predictions, and Cov denotes acceptance coverage. 
  Agreement filter denotes that only consistent results are accepted. VLM SMILES revision used GPT-5.6-sol.
  }
  \vspace{-1em}
  \label{tab:mol_info}

  \renewcommand\arraystretch{0.95}
  \setlength{\tabcolsep}{7pt}

  \newcommand{\dd}[1]{%
    \rlap{\hspace{0.08em}{%
      \raisebox{0.55ex}{%
        \fontsize{4.2}{4.2}\selectfont
        \textcolor{ForestGreen}{+#1}%
      }%
    }}%
  }

  \sisetup{
    table-number-alignment=center,
    detect-weight=true,
    detect-inline-weight=math
  }

  \begin{tabular*}{\textwidth}{
    @{\extracolsep{\fill}}
    ll
    S[table-format=2.2]
    S[table-format=2.2]
    S[table-format=3.2]
    @{\hspace{7pt}}
    S[table-format=2.2]
    S[table-format=2.2]
    S[table-format=3.2]
    @{}
  }

    \toprule
    \multirow{2.5}{*}{\textbf{Category}}
      & \multirow{2.5}{*}{\textbf{Method}}
      & \multicolumn{3}{c}{\textbf{OCIE}}
      & \multicolumn{3}{c}{\textbf{JIE}} \\

    \cmidrule(lr){3-5}
    \cmidrule(lr){6-8}

      &
      & {OA$\uparrow$}
      & {AA$\uparrow$}
      & {Cov$\uparrow$}
      & {OA$\uparrow$}
      & {AA$\uparrow$}
      & {Cov$\uparrow$} \\

    \midrule

    \multirow{3}{*}{\textit{VLM (e2e)}}
        & Qwen2.5-VL-3B~\citep{bai2025qwen2}
      & 0.54 & 0.54 & 100.0
      & 0.90 & 0.90 & 100.0 \\
    
      & GPT-4o~\citep{openai2026api}
      & 13.91 & 13.91 & 100.0
      & 23.58 & 23.58 & 100.0 \\

    & GPT-5.6-sol~\citep{openai2026api}
      & 49.04 & 49.04 & 100.0
      & 51.95 & 51.95 & 100.0 \\

    \midrule

    \multirow{5}{*}{\textit{Expert}}
      & OSRA~\citep{filippov2009optical}
      & 36.17 & 36.17 & 100.0
      & 19.88 & 19.88 & 100.0 \\

      & DECIMER~\citep{rajan2020decimer}
      & 35.07 & 35.07 & 100.0
      & 16.42 & 16.42 & 100.0 \\

      & MolGrapher~\citep{morin2023molgrapher}
      & 41.20 & 41.20 & 100.0
      & 25.40 & 25.40 & 100.0 \\

      & MolNexTR~\citep{chen2024molnextr}
      & 73.24 & 73.24 & 100.0
      & 61.62 & 61.62 & 100.0 \\

      & MolScribe~\citep{qian2023molscribe}
      & 73.53 & 73.53 & 100.0
      & 61.83 & 61.83 & 100.0 \\

    \midrule

    \multirow{4}{*}{\shortstack{\textit{SSR (with}\\\textit{MolScribe)}}}

      & Confidence filter (AA-matched)
      & 73.53 & 97.50 & 52.3
      & 61.83 & 97.08 & 36.7 \\

      & Confidence filter (Cov-matched)
        & 73.53 & 80.12 & 89.0
      & 61.83 & 67.48 & 85.4 \\

      & Agreement filter with MolNexTR
      & 73.53 & 91.83 & 74.3
      & 61.83 & 88.74 & 61.6 \\

      & VLM SMILES revision
      & 78.91 & 78.91 & 100.0
      & 68.74 & 68.74 & 100.0 \\

    \midrule

    \multirow{3}{*}{\textbf{Ours}}
        & MolGrapher + ViCoR
      & {76.84\dd{35.64}}
      & {95.68\dd{54.48}}
      & 78.2
      & {66.95\dd{41.55}}
      & {94.81\dd{69.41}}
      & 69.8 \\
    
      & MolNexTR + ViCoR
      & {\underline{88.08}\dd{14.84}}
      & {\underline{97.44}\dd{24.20}}
      & 88.9
      & {\bfseries 84.32\dd{22.70}}
      & {\underline{97.10}\dd{35.48}}
      & 85.3 \\

      & \textbf{MolScribe + ViCoR}
      & {\bfseries 88.26\dd{14.73}}
      & {\bfseries 97.54\dd{24.01}}
      & 89.0
      & {\bfseries 84.32\dd{22.49}}
      & {\bfseries 97.17\dd{35.34}}
      & 85.4 \\

    \bottomrule
  \end{tabular*}

  \vspace{-1em}
\end{table*}

\begin{table*}[t]
  \footnotesize
  \centering
  \caption{Exact-match accuracy (OA) on eight standard OCSR benchmarks.
  $\Delta$ denotes the gain of ViCoR over its base recognizer, MolScribe.
  }
  \vspace{-1em}
  \label{tab:public}

  \renewcommand\arraystretch{0.82}
  \setlength{\tabcolsep}{2pt}

  \newcommand{\dd}[1]{%
    \textcolor{ForestGreen}{\scriptsize +#1}%
  }

  \sisetup{
    table-number-alignment=center,
    table-format=2.1,
    detect-weight=true,
    detect-inline-weight=math
  }

  \resizebox{\textwidth}{!}{%
  \begin{tabular*}{\textwidth}{
    @{\extracolsep{\fill}}
    ll
    S S
    @{\hspace{7pt}}
    S S S S S S
    @{}
  }
    \toprule

    \multirow{2}{*}{\textbf{Category}}
      & \multirow{2}{*}{\textbf{Method}}
      & \multicolumn{2}{c}{\textbf{Synthetic}}
      & \multicolumn{6}{c}{\textbf{Realistic}} \\

    \cmidrule(lr){3-4}
    \cmidrule(lr){5-10}

      &
      & {Indigo} & {ChemDraw}
      & {CLEF} & {UOB} & {JPO} & {USPTO} & {Staker} & {ACS} \\

    \midrule

    \multirow{2}{*}{\textit{Rule-based}}
      & MolVec~\citep{peryea2019molvec}
      & 95.4 & 87.9
      & 82.8 & 80.6 & 67.8 & 88.4 & 0.8 & 47.4 \\

      & OSRA~\citep{filippov2009optical}
      & 95.0 & 87.3
      & 84.6 & 78.5 & 55.3 & 87.4 & 0.0 & 55.3 \\

    \midrule

    \multirow{3}{*}{\textit{End-to-end}}

      & DECIMER~\citep{rajan2020decimer}
      & 69.6 & 86.1
      & 62.7 & 88.2 & 55.2 & 41.1 & 40.8 & 46.5 \\

      & MolParser~\citep{fang2025molparser}
      & \multicolumn{1}{c}{---}
      & \multicolumn{1}{c}{---}
      & 91.0 & 91.6 & 75.6 & 93.0
      & \multicolumn{1}{c}{---}
      & \multicolumn{1}{c}{---} \\

      & MolSight~\citep{zhang2026molsight}
      & \multicolumn{1}{c}{---}
      & \multicolumn{1}{c}{---}
      & 85.5 & 87.4 & 57.6 & 92.0
      & \multicolumn{1}{c}{---}
      & \multicolumn{1}{c}{---} \\

    \midrule

    \multirow{3}{*}{\textit{Graph-based}}

      & MolGrapher~\citep{morin2023molgrapher}
      & \multicolumn{1}{c}{---}
      & \multicolumn{1}{c}{---}
      & 90.5
      & \multicolumn{1}{c}{---}
      & 67.5 & 91.5
      & \multicolumn{1}{c}{---}
      & \multicolumn{1}{c}{---} \\

      & MolNexTR~\citep{chen2024molnextr}
      & 97.8 & 95.1
      & 90.4 & 88.5 & 82.1 & 93.8 & 88.3 & 81.9 \\

      & MolScribe~\citep{qian2023molscribe}
      & 97.5 & 93.8
      & 88.9 & 87.4 & 76.2 & 93.1 & 86.9 & 71.9 \\

    \midrule

    \multirow{2}{*}{\textbf{Ours}}
      & \textbf{MolScribe + ViCoR}
      & {\bfseries 98.5}
      & {\bfseries 96.4}
      & {\bfseries 93.8}
      & {\bfseries 92.8}
      & {\bfseries 86.5}
      & {\bfseries 94.7}
      & {\bfseries 90.7}
      & {\bfseries 89.7} \\

    \cmidrule(l){2-10}

      & \textit{$\Delta$ vs.\ MolScribe}
      & \multicolumn{1}{c}{\dd{1.0}}
      & \multicolumn{1}{c}{\dd{2.6}}
      & \multicolumn{1}{c}{\dd{4.9}}
      & \multicolumn{1}{c}{\dd{5.4}}
      & \multicolumn{1}{c}{\dd{10.3}}
      & \multicolumn{1}{c}{\dd{1.6}}
      & \multicolumn{1}{c}{\dd{3.8}}
      & \multicolumn{1}{c}{\dd{17.8}} \\

    \bottomrule
  \end{tabular*}%
  }

\end{table*}

\subsection{Selective Structure Recognition for Reliable OCSR}

Table~\ref{tab:mol_info} compares ViCoR with end-to-end VLMs (e.g., GPT-5.6-sol), expert OCSR models (e.g., DECIMER), and baseline SSR strategies (e.g., confidence filtering) on OCIE and JIE.
For VLM SMILES revision, GPT-5.6-sol directly revises the MolScribe prediction conditioned on the source image.
ViCoR consistently improves both OA and AA across different graph-based recognizers.

\textbf{Observation 1: ViCoR improves recognition beyond filtering.}
With MolScribe as the base recognizer, ViCoR raises OA from 73.53 to 88.26 on OCIE and from 61.83 to 84.32 on JIE.
In contrast, selection-only methods leave OA unchanged because they do not modify the initial predictions.
Direct VLM SMILES revision improves OA to 78.91 and 68.74, but remains well below ViCoR, highlighting the difficulty of revising structures directly across image and symbolic representations.
By establishing spatial correspondence and executable graph edits, ViCoR can instead revise localized recognition errors while preserving the
remaining structure. Table~\ref{tab:public} further shows consistent OA improvements on eight standard OCSR benchmarks, such as JPO (+10.3) and ACS (+17.8).

\textbf{Observation 2: ViCoR achieves high reliability at broad coverage.}
ViCoR achieves 97.54\%/97.17\% AA at 89.0\%/85.4\% coverage on OCIE/JIE,
yielding 86.9\%/83.0\% of all predictions that are both accepted and correct,
already exceeding the base recognizer's OA.
Compared with confidence filtering, ViCoR achieves higher AA at matched
coverage and higher coverage at matched AA.
Agreement filtering also reaches high AA but with substantially lower coverage,
since disagreements can only be rejected rather than revised.
Overall, ViCoR combines reliable acceptance with broad coverage, which further
benefits downstream chemical tasks in Sec.~\ref{sec:application}.

\vspace{-0.5em}
\subsection{Real-world Downstream Utility of Reliable Structures}\label{sec:application}
\vspace{-0.5em}

\textbf{Chemical Reaction Extraction from Literature.}
Automated reaction extraction is a complex, multi-stage pipeline (e.g., template and product recognition, and R-group substitution), highly sensitive to single errors. 
We integrate ViCoR into OpenChemIE~\citep{fan2024openchemie} and evaluate
reaction-level extraction performance in Tab.~\ref{tab:openchemie}.
Confidence filtering improves reliability by rejecting uncertain structures,
but substantially reduces recall and overall F1.
VLM SMILES revision provides a moderate improvement over raw OCSR outputs.
In contrast, ViCoR substantially improves both precision and F1, showing that reliable structure recognition directly benefits
downstream extraction.

\begin{table*}[t]
  \footnotesize
  \centering
  
  \begin{minipage}{0.3\textwidth}
    \centering
    \scriptsize
    \caption{Chemical reaction extraction performance using OpenChemIE tool.}
    \vspace{-1em}
    \label{tab:openchemie}
    \renewcommand\arraystretch{1.1}
    \setlength{\tabcolsep}{2pt}
    \begin{tabular}{@{}lccc@{}}
      \toprule
      \textbf{Methods} & \textbf{Precision} & \textbf{Recall} & \textbf{F1} \\
      \midrule
      Raw OCSR      & 80.85 & 51.14 & 62.65 \\
      confidence filter & 80.70 & 36.54 & 50.31 \\
      Agreement filter & 85.10 & 44.70 & 58.61 \\
      VLM SMILES rev. & 85.60 & 57.80 & 69.01 \\
      \midrule
      \textbf{ViCoR (Verify only)}
      & 92.30 & 48.90 & 63.93 \\
      \textbf{ViCoR (Rev. only)}
      & 90.85 & 62.80 & 74.26 \\
      \textbf{ViCoR}     & \textbf{95.72} & \textbf{65.96} & \textbf{78.10} \\
      \bottomrule
    \end{tabular}
  \end{minipage}
  \hfill
  \begin{minipage}{0.65\textwidth}
    \centering
    \scriptsize
    \caption{Forward reaction prediction performance using literature-sourced training data from different SSR strategies.}
    \vspace{-1em}
    \label{tab:reaction_prediction}
    \renewcommand\arraystretch{1.1}
    \setlength{\tabcolsep}{1.6pt}

    \begin{tabular}{lccc@{\hspace{8pt}}ccc}
      \toprule
      \multirow{2}{*}{\textbf{Methods}}
      & \multicolumn{3}{c}{\textbf{Suzuki Coupling}}
      & \multicolumn{3}{c}{\textbf{Photocatalysis}} \\
      \cmidrule(lr){2-4}
      \cmidrule(lr){5-7}
      & EXACT$\uparrow$
      & RDK FTS$\uparrow$
      & Morgan$\uparrow$
      & EXACT$\uparrow$
      & RDK FTS$\uparrow$
      & Morgan$\uparrow$ \\
      \midrule

      Raw OCSR
      & 0.207 & 0.445 & 0.484
      & 0.173 & 0.471 & 0.480 \\

    Confidence filtering
      & 0.212 & 0.440 & 0.486
      & 0.189 & 0.486 & 0.492 \\

    Agreement filtering
      & 0.218 & 0.448 & 0.490
      & 0.195 & 0.489 & 0.496 \\

      VLM SMILES rev.
      & 0.234 & 0.462 & 0.496
      & 0.198 & 0.496 & 0.499 \\
      
      \midrule

        \textbf{ViCoR (Verify only)}
      & 0.220
      & 0.444
      & 0.491
      & 0.201
      & 0.494
      & 0.502 \\

        \textbf{ViCoR (Rev. only)}
      & 0.265
      & 0.487
      & 0.523
      & 0.211
      & 0.508
      & 0.509 \\

      \textbf{ViCoR}
      & \textbf{0.284}
      & \textbf{0.503}
      & \textbf{0.536}
      & \textbf{0.231}
      & \textbf{0.521}
      & \textbf{0.526} \\

      \bottomrule
    \end{tabular}
  \end{minipage}

  \vspace{-1em}
\end{table*}

\begin{table*}[t]
  \footnotesize
  \centering
  \caption{
  Ablation of the spatially aligned visual comparison.
  \textit{Spat. Align} denotes coordinate-preserving rendering,
  \textit{SBS} side-by-side presentation, and
  \textit{Index} explicit atom indexing.
  }
  \vspace{-0.8em}
  \label{tab:representation}

  \renewcommand{\arraystretch}{1.0}
  \setlength{\tabcolsep}{8pt}

  \resizebox{\textwidth}{!}{%
  \begin{tabular}{
    @{}
    l ccc
    rrr
    rrr
    @{}
  }
    \toprule
    \multirow{2}{*}{\textbf{Interface}}
      & \multicolumn{3}{c}{\textbf{Design}}
      & \multicolumn{3}{c}{\textbf{OCIE}}
      & \multicolumn{3}{c}{\textbf{JIE}} \\
    \cmidrule(lr){2-4}
    \cmidrule(lr){5-7}
    \cmidrule(lr){8-10}
      & Spat. Align & SBS & Index
      & OA$\uparrow$ & AA$\uparrow$ & Cov$\uparrow$
      & OA$\uparrow$ & AA$\uparrow$ & Cov$\uparrow$ \\
    \midrule

    Base recognizer
      & -- & -- & --
      & 73.53 & 73.53 & 100.0
      & 61.83 & 61.83 & 100.0 \\

    \midrule
    \multicolumn{10}{@{}l}{\textit{Source-prediction interfaces}} \\

    \quad Image \& SMILES
      & \xmark & \xmark & \xmark
      & 74.26 & 82.70 & 68.8
      & 62.91 & 77.80 & 59.6 \\

    \quad Image \& graph
      & \xmark & \xmark & \cmark
      & 76.48 & 86.90 & 73.4
      & 65.74 & 81.90 & 65.1 \\

    \midrule
    \multicolumn{10}{@{}l}{\textit{Rendered interfaces}} \\

    \quad Overlay
      & \cmark & \xmark & \cmark
      & 83.64 & 93.60 & 84.2
      & 78.42 & 91.40 & 78.3 \\

    \quad Side-by-side, default layout
      & \xmark & \cmark & \cmark
      & 79.71 & 89.70 & 78.1
      & 72.88 & 86.20 & 71.5 \\

    \quad Side-by-side, coord.-preserving
      & \cmark & \cmark & \xmark
      & 82.85 & 96.84 & 84.7
      & 77.81 & 96.35 & 80.1 \\

    \quad \textbf{ViCoR (ours)}
      & \cmark & \cmark & \cmark
      & \textbf{88.26} & \textbf{97.54} & \textbf{89.0}
      & \textbf{84.32} & \textbf{97.17} & \textbf{85.4} \\

    \bottomrule
  \end{tabular}%
  }

  \vspace{-1em}
\end{table*}

\textbf{Reaction Prediction with Literature-Extracted Data.}
To further study whether SSR improved 
downstream model learning, 
we collect 2,100 photocatalysis and 430 Suzuki-coupling figures from chemical
journals and use OpenChemIE to extract reaction data under different post-recognition strategies.
The resulting datasets are then used to fine-tune Qwen3-8B~\citep{yang2025qwen3}
for forward reaction prediction.
As shown in Tab.~\ref{tab:reaction_prediction}, confidence filtering provides
only modest gains because cleaner data comes at the cost of reduced training
coverage, while VLM SMILES revision recovers part of the corrupted data.
ViCoR achieves the best performance across both reaction domains, 
showing that improved structure reliability and
coverage translate into better downstream learning.

\vspace{-0.5em}
\subsection{Ablation and Design Analysis}
\vspace{-0.5em}

\textbf{Spatially Aligned Visual Comparison.}
Table~\ref{tab:representation} studies how the prediction interface affects
verification and revision.
Directly comparing the source image with symbolic representations such as
SMILES or graph performs poorly because the model must infer the
correspondence between visual regions and symbolic entities.
Rendering the prediction reduces this representation gap, but canonical-layout
rendering still changes the spatial arrangement of atoms and bonds.
In contrast, coordinate-preserving rendering substantially improves both OA
and AA by explicitly aligning corresponding regions.
Side-by-side presentation provides a further gain over image overlay, while
atom indices further improve performance by making graph entities directly
addressable for revision.
Overall, these results validate spatial alignment and index anchoring as the key interfaces.

\textbf{Iterative Revision and Computational Cost.}
Tab.~\ref{tab:ablation iteration} evaluates different revision budgets.
Increasing the number of iterations consistently improves OA, showing that 
single revision may leave residual structural errors.
Re-verifying each intermediate prediction allows ViCoR to identify and revise
these remaining errors, with only a modest reduction in throughput.
Fig.~\ref{fig:examples}(b) shows an example.

\textbf{Effect of Progressive Training.}
Tab.~\ref{tab:ablation training} shows that progressive training consistently outperforms separate and mixed training.
Verification-oriented pretraining provides a stronger initialization for subsequent graph revision by learning robust visual correspondence.

\vspace{-0.5em}
\subsection{Further Analysis}
\vspace{-1em}

\begin{figure*}[t]
    \centering
    \includegraphics[width=1.0\textwidth]{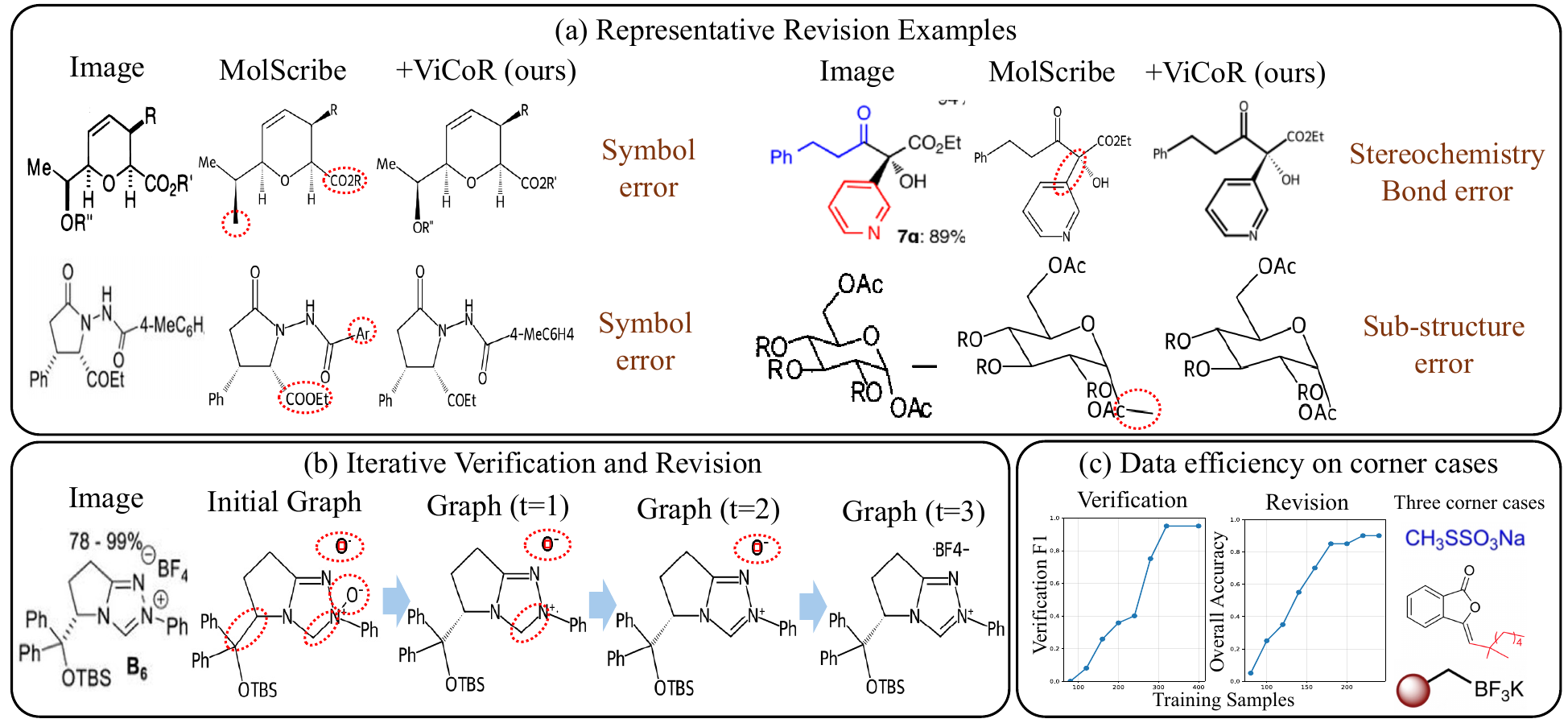}
    \vspace{-2 em}
    \caption{Qualitative results of ViCoR. (a) Representative Revision Examples. (b) ViCoR improves recognition through iterative verification and revision. (c) Data efficiency on corner cases.}
    \label{fig:examples}
    \vspace{-1 em}
\end{figure*}

\textbf{Data Efficiency on Corner Cases.}
We evaluate how efficiently ViCoR adapts to three challenging corner cases:
single-string labels, Markush repetition, and special R-group symbols
(Fig.~\ref{fig:examples}(c)).
With only 300 targeted synthetic samples, ViCoR already achieves strong
verification and revision performance.
This shows that rare failure modes can be learned with little task-specific data.

\textbf{Transfer to Another Image-to-Structure Task.}
We transfer ViCoR to hand-drawn BPMN process-graph recognition~\citep{schafer2022sketch2process}.
With Faster R-CNN~\citep{ren2016faster} and YOLO11m \citep{khanam2024yolov11}, ViCoR improves OA from 41.6\% to 62.1\% and from 50.4\% to 68.8\%, respectively, while retaining over 95\% AA, showing applicability beyond OCSR. 
Details in Appendix.

\begin{table*}[t]
  \centering

  \begin{minipage}{0.44\textwidth}
    \footnotesize
    \centering
    \caption{Ablation on iterative revision.}
    \vspace{-1 em}
    \label{tab:ablation iteration}
    \renewcommand\arraystretch{1.1}
    \setlength{\tabcolsep}{4pt} 
    \begin{tabular}{@{}lccc@{}}
      \toprule
      Models 
      & \shortstack{Thpt.(img/s)} 
      & \shortstack{OCIE(OA)} 
      & \shortstack{JIE(OA)} \\
      \midrule
      Baseline      
      & 1.61 & 73.53 & 61.83 \\
      \midrule
      $T=1$ 
      & 1.33 & 80.82 & 72.45 \\
      
      $T=2$ 
      & 1.21 & \underline{86.45} & \underline{81.36} \\
      
      $T=3$ 
      & 1.18 & \textbf{88.26} & \textbf{84.32} \\
      \bottomrule
    \end{tabular}
  \end{minipage}
  \hfill
  \begin{minipage}{0.52\textwidth}
    \footnotesize
    \centering
    \caption{Ablation on training strategy.}
    \vspace{-1 em}
    \label{tab:ablation training}
    \renewcommand\arraystretch{1.1}
    \setlength{\tabcolsep}{3pt}
    \begin{tabular}{@{}lcccccc@{}}
      \toprule
      \multirow{2}{*}{Training}
      & \multicolumn{3}{c}{OCIE}
      & \multicolumn{3}{c}{JIE} \\
      \cmidrule(lr){2-4}
      \cmidrule(lr){5-7}
      & OA$\uparrow$
      & AA$\uparrow$
      & Cov$\uparrow$
      & OA$\uparrow$
      & AA$\uparrow$
      & Cov$\uparrow$ \\
      \midrule

    Separate
      & 87.81 & 97.50 & 88.55
      & 84.04 & 97.14 & 85.11 \\

      Mixed
      & 87.45
      & 96.85
      & 88.67
      & 83.85
      & 96.95
      & 84.70 \\

      Progressive
      & \textbf{88.26}
      & \textbf{97.54}
      & \textbf{89.04}
      & \textbf{84.32}
      & \textbf{97.17}
      & \textbf{85.44} \\

      \bottomrule
    \end{tabular}
  \end{minipage}

  \vspace{-1em}
\end{table*}

\vspace{-0.5em}
\section{Conclusion}
\vspace{-1em}
We introduce \textbf{ViCoR}, a post-recognition framework for reliable OCSR
that follows a \emph{repair-before-rejection} principle.
By combining spatially aligned verification, index-anchored executable
revision, and verification-to-revision progressive training, ViCoR improves
both recognition accuracy and the reliability--coverage trade-off, while
providing higher-quality molecular data for downstream reaction extraction and
prediction.
ViCoR still has limitations on highly complex molecular depictions, especially
3D structures that are difficult to interpret from 2D images, and its iterative
inference introduces additional computational cost.
Future work will improve recognition and revision for complex 3D molecules and
explore more efficient inference for large-scale scientific data curation.

\subsection*{AI use statement}

In this work, we used generative AI tools for language polishing and improving the clarity and presentation of the manuscript. We have not used generative AI tools to formulate research hypotheses, design the methodology or experiments, or interpret experimental results, and the remaining required disclosure tasks are not applicable to this work. We have reviewed all AI-assisted edits and revised them where necessary to ensure accuracy and consistency with the authors' intended meaning. We take responsibility for the final content of this work, including text, claims, or artifacts produced with the aid of generative AI.

\subsection*{Ethics statement}

We anticipate that ViCoR can facilitate the reliable extraction and digitization of molecular structures from scientific literature, thereby supporting chemical data curation and downstream cheminformatics research. At the same time, incorrectly recognized or revised structures may propagate errors into chemical databases or downstream models if used without appropriate validation. ViCoR is intended to improve the reliability of automated structure recognition rather than replace expert verification in safety-critical applications. Our study does not involve human subjects or personal data. We encourage responsible use of the released models, code, and benchmarks in accordance with applicable licenses and research ethics guidelines.

\subsection*{Reproducibility statement}

To ensure the reproducibility of our research, we provide the implementation and experimental details of ViCoR in Section~4.1 and the Appendix, including model configurations, training settings, dataset construction, and evaluation protocols. To facilitate future research and technical transparency, we will publicly release our source code, model configurations, trained model weights, and curated benchmarks in accordance with relevant licenses. These resources will enable researchers to reproduce our results and build upon ViCoR.

\bibliography{iclr2027_conference}

@article{qian2023molscribe,
  title={MolScribe: robust molecular structure recognition with image-to-graph generation},
  author={Qian, Yujie and Guo, Jiang and Tu, Zhengkai and Li, Zhening and Coley, Connor W and Barzilay, Regina},
  journal={Journal of chemical information and modeling},
  volume={63},
  number={7},
  pages={1925--1934},
  year={2023},
  publisher={ACS Publications}
}

@article{chen2024molnextr,
  title={MolNexTR: a generalized deep learning model for molecular image recognition},
  author={Chen, Yufan and Leung, Ching Ting and Huang, Yong and Sun, Jianwei and Chen, Hao and Gao, Hanyu},
  journal={Journal of Cheminformatics},
  volume={16},
  number={1},
  pages={141},
  year={2024},
  publisher={Springer}
}

@article{weininger1988smiles,
  title={SMILES, a chemical language and information system. 1. Introduction to methodology and encoding rules},
  author={Weininger, David},
  journal={Journal of chemical information and computer sciences},
  year={1988},
  publisher={ACS Publications}
}

@article{madaan2023self,
  title={Self-refine: Iterative refinement with self-feedback},
  author={Madaan, Aman and Tandon, Niket and Gupta, Prakhar and Hallinan, Skyler and Gao, Luyu and Wiegreffe, Sarah and Alon, Uri and Dziri, Nouha and Prabhumoye, Shrimai and Yang, Yiming and others},
  journal={Advances in Neural Information Processing Systems},
  volume={36},
  pages={46534--46594},
  year={2023}
}

@article{chen2025multi,
  title={A Multi-Agent System Enables Versatile Information Extraction from the Chemical Literature},
  author={Chen, Yufan and Leung, Ching Ting and Yu, Bowen and Sun, Jianwei and Huang, Yong and Li, Linyan and Chen, Hao and Gao, Hanyu},
  journal={arXiv preprint arXiv:2507.20230},
  year={2025}
}

@article{kim2016pubchem,
  title={PubChem substance and compound databases},
  author={Kim, Sunghwan and Thiessen, Paul A and Bolton, Evan E and Chen, Jie and Fu, Gang and Gindulyte, Asta and Han, Lianyi and He, Jane and He, Siqian and Shoemaker, Benjamin A and others},
  journal={Nucleic acids research},
  volume={44},
  number={D1},
  pages={D1202--D1213},
  year={2016},
  publisher={Oxford University Press}
}

@misc{filippov2009optical,
  title={Optical structure recognition software to recover chemical information: OSRA, an open source solution},
  author={Filippov, Igor V and Nicklaus, Marc C and Nicklaus, Marc C},
  year={2009},
  publisher={ACS Publications}
}

@inproceedings{piroi2010clef,
  title={CLEF-IP 2010: Retrieval Experiments in the Intellectual Property Domain.},
  author={Piroi, Florina and Lupu, Mihai and Hanbury, Allan and Sexton, Alan P and Magdy, Walid and Filippov, Igor V},
  booktitle={CLEF (notebook papers/labs/workshops)},
  year={2010}
}

@inproceedings{fujiyoshi2011robust,
  title={Robust method of segmentation and recognition of chemical structure images in cheminfty},
  author={Fujiyoshi, Akio and Nakagawa, Koji and Suzuki, Masakazu},
  booktitle={Pre-proceedings of the 9th IAPR international workshop on graphics recognition, GREC},
  volume={1},
  year={2011}
}

@inproceedings{sadawi2012chemical,
  title={Chemical structure recognition: a rule-based approach},
  author={Sadawi, Noureddin M and Sexton, Alan P and Sorge, Volker},
  booktitle={Document recognition and retrieval XIX},
  pages={101--109},
  year={2012},
  organization={SPIE}
}

@article{staker2019molecular,
  title={Molecular structure extraction from documents using deep learning},
  author={Staker, Joshua and Marshall, Kyle and Abel, Robert and McQuaw, Carolyn M},
  journal={Journal of chemical information and modeling},
  year={2019},
  publisher={ACS Publications}
}

@article{fan2024openchemie,
  title={OpenChemIE: An information extraction toolkit for chemistry literature},
  author={Fan, Vincent and Qian, Yujie and Wang, Alex and Wang, Amber and Coley, Connor W and Barzilay, Regina},
  journal={Journal of Chemical Information and Modeling},
  volume={64},
  number={14},
  pages={5521--5534},
  year={2024},
  publisher={ACS Publications}
}

@article{hu2022lora,
  title={Lora: Low-rank adaptation of large language models.},
  author={Hu, Edward J and Shen, Yelong and Wallis, Phillip and Allen-Zhu, Zeyuan and Li, Yuanzhi and Wang, Shean and Wang, Lu and Chen, Weizhu and others},
  journal={ICLR},
  volume={1},
  number={2},
  pages={3},
  year={2022}
}

@article{bai2025qwen2,
  title={Qwen2. 5-vl technical report},
  author={Bai, Shuai and Chen, Keqin and Liu, Xuejing and Wang, Jialin and Ge, Wenbin and Song, Sibo and Dang, Kai and Wang, Peng and Wang, Shijie and Tang, Jun and others},
  journal={arXiv preprint arXiv:2502.13923},
  year={2025}
}

@inproceedings{peryea2019molvec,
  title={MOLVEC: Open source library for chemical structure recognition},
  author={Peryea, Tyler and Katzel, Daniel and Zhao, Tongan and Southall, Noel and Nguyen, Dac-Trung},
  booktitle={Abstracts of papers of the American Chemical Society},
  year={2019},
  organization={Amer Chemical Soc 1155 16TH ST, NW, WASHINGTON, DC 20036 USA}
}

@article{rajan2020decimer,
  title={DECIMER: towards deep learning for chemical image recognition},
  author={Rajan, Kohulan and Zielesny, Achim and Steinbeck, Christoph},
  journal={Journal of Cheminformatics},
  volume={12},
  number={1},
  pages={65},
  year={2020},
  publisher={Springer}
}

@inproceedings{fang2025molparser,
  title={Molparser: End-to-end visual recognition of molecule structures in the wild},
  author={Fang, Xi and Wang, Jiankun and Cai, Xiaochen and Chen, Shangqian and Yang, Shuwen and Tao, Haoyi and Wang, Nan and Yao, Lin and Zhang, Linfeng and Ke, Guolin},
  booktitle={Proceedings of the IEEE/CVF International Conference on Computer Vision},
  pages={24528--24538},
  year={2025}
}

@inproceedings{morin2023molgrapher,
  title={MolGrapher: graph-based visual recognition of chemical structures},
  author={Morin, Lucas and Danelljan, Martin and Agea, Maria Isabel and Nassar, Ahmed and Weber, Valery and Meijer, Ingmar and Staar, Peter and Yu, Fisher},
  booktitle={Proceedings of the IEEE/CVF International Conference on Computer Vision},
  year={2023}
}

@article{chen2025towards,
  title={Towards Large-scale Chemical Reaction Image Parsing via a Multimodal Large Language Model},
  author={Chen, Yufan and Leung, Ching Ting and Sun, Jianwei and Huang, Yong and Li, Linyan and Chen, Hao and Gao, Hanyu},
  journal={arXiv preprint arXiv:2503.08156},
  year={2025}
}

@article{song2025rxncaption,
  title={RxnCaption: Reformulating Reaction Diagram Parsing as Visual Prompt Guided Captioning},
  author={Song, Jiahe and Wang, Chuang and Jiang, Bowen and Wang, Yinfan and Zheng, Hao and Wei, Xingjian and Liu, Chengjin and Gao, Junyuan and Wang, Yubin and Wu, Lijun and others},
  journal={arXiv preprint arXiv:2511.02384},
  year={2025}
}

@article{leong2025mermaid,
  title={MERMaid: Universal multimodal mining of chemical reactions from PDFs using vision-language models},
  author={Leong, Shi Xuan and Pablo-Garc{\'\i}a, Sergio and Wong, Brandon and Aspuru-Guzik, Al{\'a}n},
  year={2025}
}

@misc{rdkit2016,
  author       = {Landrum, Greg},
  title        = {RDKit: Open-Source Cheminformatics Software},
  year         = {2016},
  howpublished = {\url{https://www.rdkit.org}},
  note         = {Accessed 2016}
}

@inproceedings{zhang2026molsight,
  title={MolSight: Optical Chemical Structure Recognition with SMILES Pretraining, Multi-Granularity Learning and Reinforcement Learning},
  author={Zhang, Wenrui and Wang, Xinggang and Feng, Bin and Liu, Wenyu},
  booktitle={Proceedings of the AAAI Conference on Artificial Intelligence},
  year={2026}
}

@article{morin2024patcid,
  title={PatCID: an open-access dataset of chemical structures in patent documents},
  author={Morin, Lucas and Weber, Val{\'e}ry and Meijer, Gerhard Ingmar and Yu, Fisher and Staar, Peter WJ},
  journal={Nature Communications},
  volume={15},
  number={1},
  pages={6532},
  year={2024},
  publisher={Nature Publishing Group UK London}
}

@article{yang2025qwen3,
  title={Qwen3 technical report},
  author={Yang, An and Li, Anfeng and Yang, Baosong and Zhang, Beichen and Hui, Binyuan and Zheng, Bo and Yu, Bowen and Gao, Chang and Huang, Chengen and Lv, Chenxu and others},
  journal={arXiv preprint arXiv:2505.09388},
  year={2025}
}

@misc{openai2026api,
  author = {{OpenAI}},
  title = {OpenAI API},
  howpublished = {\url{https://openai.com/api/}},
  year = {2026},
  note = {Accessed: 2026-05-03}
}

@article{geifman2017selective,
  title={Selective classification for deep neural networks},
  author={Geifman, Yonatan and El-Yaniv, Ran},
  journal={Advances in neural information processing systems},
  volume={30},
  year={2017}
}

@article{tan2026data,
  title={Data-Driven, Mechanistically Guided Prediction of Yield and Chemoselectivity in SuFEx Reactions},
  author={Tan, Hao-Dong and Gao, Ben and Huang, Huaihai and Xu, Tingjun and Li, Yao and Dong, Jiajia and Xue, Xiao-Song},
  journal={Journal of the American Chemical Society},
  volume={148},
  number={23},
  pages={24138--24150},
  year={2026},
  publisher={ACS Publications}
}

@article{nippa2024enabling,
  title={Enabling late-stage drug diversification by high-throughput experimentation with geometric deep learning},
  author={Nippa, David F and Atz, Kenneth and Hohler, Remo and M{\"u}ller, Alex T and Marx, Andreas and Bartelmus, Christian and Wuitschik, Georg and Marzuoli, Irene and Jost, Vera and Wolfard, Jens and others},
  journal={Nature Chemistry},
  volume={16},
  number={2},
  pages={239--248},
  year={2024},
  publisher={Nature Publishing Group UK London}
}

@article{el2010foundations,
  title={On the Foundations of Noise-free Selective Classification.},
  author={El-Yaniv, Ran and others},
  journal={Journal of Machine Learning Research},
  volume={11},
  number={5},
  year={2010}
}

@inproceedings{geifman2019selectivenet,
  title={Selectivenet: A deep neural network with an integrated reject option},
  author={Geifman, Yonatan and El-Yaniv, Ran},
  booktitle={International conference on machine learning},
  pages={2151--2159},
  year={2019},
  organization={PMLR}
}

@inproceedings{gou2024critic,
  title={Critic: Large language models can self-correct with tool-interactive critiquing},
  author={Gou, Zhibin and Shao, Zhihong and Gong, Yeyun and Yang, Yujiu and Duan, Nan and Chen, Weizhu and others},
  booktitle={International Conference on Learning Representations},
  volume={2024},
  pages={57734--57811},
  year={2024}
}

@inproceedings{srinivasan2024selective,
  title={Selective “selective prediction”: Reducing unnecessary abstention in vision-language reasoning},
  author={Srinivasan, Tejas and Hessel, Jack and Gupta, Tanmay and Lin, Bill Yuchen and Choi, Yejin and Thomason, Jesse and Chandu, Khyathi},
  booktitle={ACL 2024 Finding},
  pages={12935--12948},
  year={2024}
}

@article{ren2016faster,
  title={Faster R-CNN: Towards real-time object detection with region proposal networks},
  author={Ren, Shaoqing and He, Kaiming and Girshick, Ross and Sun, Jian},
  journal={IEEE transactions on pattern analysis and machine intelligence},
  volume={39},
  number={6},
  pages={1137--1149},
  year={2016},
  publisher={IEEE}
}

@article{khanam2024yolov11,
  title={Yolov11: An overview of the key architectural enhancements},
  author={Khanam, Rahima and Hussain, Muhammad and Hussain, Muhammad},
  journal={arXiv preprint arXiv:2410.17725},
  year={2024}
}

@article{schafer2022sketch2process,
  title={Sketch2process: End-to-end BPMN sketch recognition based on neural networks},
  author={Sch{\"a}fer, Bernhard and Van Der Aa, Han and Leopold, Henrik and Stuckenschmidt, Heiner},
  journal={IEEE Transactions on Software Engineering},
  volume={49},
  number={4},
  pages={2621--2641},
  year={2022},
  publisher={IEEE}
}
\bibliographystyle{iclr2027_conference}

\appendix
\section{Training Data Synthesis}

We present the training data synthesis procedure for ViCoR. The pipeline includes a set of common data augmentations applied to both verification and revision tasks, along with task-specific synthesis methods designed to support their respective learning goals.
Finally, we introduce the data synthesis for corner cases.

\subsection{Data Augmentation}
To improve the diversity and robustness of the synthesized molecular images, we apply a combination of rendering-level, image-level, and molecular-level augmentations during data synthesis.

\textbf{Rendering-level Augmention.}
Molecular depictions in real-world chemical literature exhibit substantial stylistic variation. Following~\citep{chen2024molnextr}, we introduce randomized rendering variations using Indigo to simulate this diversity. Specifically, we vary bond thicknesses, line widths, font families and sizes, inter-line spacing for double and triple bonds, label rendering modes, and the visibility of implicit hydrogens.

\textbf{Image-level Augmentation.}
After rendering the image, 
we apply additional image-level augmentations to further diversify the appearance of the synthesized images. The applied transformations include: (1) random Gaussian blur, (2) random grayscale conversion, (3) random white padding, and (4) random insertion of background circles for individual atoms.

\textbf{Molecular-level Augmentation.}
Although the molecular SMILES sourced from PubChem~\citep{kim2016pubchem} provide substantial structural diversity, they lack sufficient coverage of R-group notation and commonly used abbreviations (e.g., 4-MeC$_6$H$_4$) frequently observed in chemical literature. To address this limitation, we construct curated lists of R-group symbols and 66 commonly used complex abbreviations (e.g., O-\textit{t}-Bu, 4-CNC$_6$H$_4$). After Indigo parses a SMILES string into a molecular graph, we randomly replace selected terminal nodes with R-group symbols or abbreviations. The modified graph is then rendered into an image by Indigo. This augmentation introduces realistic symbolic variations into the training distribution, improving the model’s robustness to R-group and abbreviation usage in real-world molecular images.

\subsection{Verification Training Data Synthesis}
To construct training data for the verification (800k samples), we describe the image rendering and concatenation procedure and the strategy used to generate negative samples. Positive samples are obtained straightforwardly by using the same molecular graph for both image renderings.

\textbf{Image Rendering and Concatenation.}
As illustrated in Fig.~\ref{fig:overview}, given a predicted molecular graph, we render a corresponding molecular image using its atomic coordinates and bond connectivity, ensuring strict spatial alignment with the original input image. The original image and the rendered image are then concatenated into a single composite image, separated by a clear black divider to visually distinguish the two views. During verification, ViCoR takes the concatenated image together with a verification prompt as input.

\textbf{Negative Sample Construction.}
Rendering an image from the ground-truth molecular graph yields only positive verification samples. To obtain informative negative examples, we employ two complementary strategies.

\emph{(1) Failure Cases from an Expert OCSR Model.}
To capture realistic error patterns produced by OCSR systems, we incorporate incorrect predictions generated by MolScribe~\citep{qian2023molscribe} as negative samples. Specifically, we synthesize 1M molecular images from PubChem SMILES using Indigo and apply MolScribe to predict their molecular graphs. The predicted graphs are then matched against the ground-truth graphs provided by Indigo using a graph isomorphism algorithm. Samples with mismatched graphs (approximately 140k) are labeled as negative examples, rendered into images, and concatenated with the corresponding original images.

\emph{(2) Random Graph Perturbation.} 
Starting from the ground-truth (GT) molecular graph corresponding to the original image, we introduce controlled random perturbations to synthesize challenging negative samples. These perturbations are systematically applied at both the atom and bond levels—specifically through the addition, substitution, and deletion of atoms and bonds—to comprehensively cover the six standard types of graph edit operations. To faithfully simulate common OCSR failures, we distribute these perturbations across three typical error categories in our training set: atom symbol errors (33\%), bond type errors (33\%), and sub-structure errors (34\%). Furthermore, to ensure varying levels of difficulty, each perturbed molecule is injected with a random number of 1 to 6 errors.

\subsection{Revision Training Data Synthesis}
To construct training data for the revision task (500k samples), we describe the image rendering and concatenation procedure and the generation of revision supervision.

\textbf{Image Rendering and Concatenation.}
The rendering process largely follows that used for the verification task. To enable location-precise graph revision, we additionally annotate each atom in the rendered image with a unique index. This is implemented by enabling the \textit{addAtomIndices} option when rendering molecular graphs with RDKit.

\textbf{Revision Format Identification.}
Unlike the verification task which relies on holistic negative samples, the revision task requires fine-grained, localized structural supervision. To construct training samples for this generative task, we apply a controlled sequence of graph edit operations to corrupt the ground-truth (GT) molecular graph. The supervision target for the VLM is then formulated as the sequence of corrective edits $\Delta$ required to restore the corrupted graph back to its original GT.

To seamlessly integrate the mathematical edit space $\Omega$ (defined in Eq.~\ref{eq:node_edits}\&~\ref{eq:edge_edits}) with the autoregressive nature of VLMs, we serialize the sequence of corrective operators into a structured, JSON-like text format. Each edit instruction $\delta_k \in \Delta$ is represented as a dictionary mapping the operation type to its specific parameters. Explicitly, the six types of graph edits are formatted as follows:

\begin{itemize}
    \item \textbf{Node Addition} ($\text{Add}_V$): \texttt{\{'Add atom': [$i$, $\ell_i$, [$x_i$, $y_i$]]\}}, where $i$ is the new node index, $\ell_i$ is the atomic symbol, and $[x_i, y_i]$ are the normalized 2D coordinates .
    \item \textbf{Node Deletion} ($\text{Del}_V$): \texttt{\{'Del atom': [$i$, $\ell_i$]\}}.
    \item \textbf{Node Revision} ($\text{Rev}_V$): \texttt{\{'Rev atom': [$i$, $\ell_i'$]\}}, denoting the correction of the atom at index $i$ to the target symbol $\ell_i'$.
    \item \textbf{Edge Addition} ($\text{Add}_E$): \texttt{\{'Add bond': [[$i$, $j$], $e_{i,j}$]\}}, specifying the source($i$) and target($j$) node indices, and the discrete bond type (e.g., \emph{solid wedge}, \emph{single}).
    \item \textbf{Edge Deletion} ($\text{Del}_E$): \texttt{\{'Del bond': [[$i$, $j$], $e_{i,j}$]\}}.
    \item \textbf{Edge Revision} ($\text{Rev}_E$): \texttt{\{'Rev bond': [[$i$, $j$], $e_{i,j}'$]\}}.
\end{itemize}

During training, an entire revision target sequence is concatenated into a single list. For instance, a complex revision sample involving multiple topological corrections is supervised using the format:
\begin{center}
\begin{minipage}{0.8\linewidth}
\ttfamily
[ \\
\hspace*{1.5em}\{'Add atom': [3, 'C', [<0.312>, <0.578>]]\}, \\
\hspace*{1.5em}\{'Add bond': [[3, 4], 'solid wedge']\}, \\
\hspace*{1.5em}\{'Del atom': [9, 'N']\}, \\
\hspace*{1.5em}\{'Del bond': [[1, 2], 'single']\}, \\
\hspace*{1.5em}\{'Rev atom': [6, 'H']\}, \\
\hspace*{1.5em}\{'Rev bond': [[4, 5], 'double']\} \\
]
\end{minipage}
\end{center}

Consistent with the VLM's tokenizer design mentioned in the main text, the continuous spatial coordinates (e.g., \texttt{0.312}, \texttt{0.578}) in the \texttt{Add atom} instruction are dynamically mapped to their corresponding quantized location tokens (e.g., \texttt{<0.312>}) during the text generation process. This structured prompt formulation effectively bridges the gap between abstract graph topology and language modeling. Fig.~\ref{fig:vlm example} illustrates an example of the input and output of the VLM.

\textbf{Training Sample Construction.}
To construct a robust training set that faithfully simulates realistic OCSR failures, we systematically control both the distribution and the topological coherence of the injected perturbations. While atom symbol errors (33\%) and bond type errors (33\%) typically involve isolated revisions ($\text{Rev}_V$ or $\text{Rev}_E$), the sub-structure errors (34\%) are designed as edits. 

Unlike naive random modifications, these composite edits consist of coordinated operations to simulate complex structural hallucinations while maintaining valid graph topologies. For instance, the addition of a hallucinated node is strictly coupled with the addition of corresponding connecting edges ($\text{Add}_V \to \text{Add}_E$) to prevent disconnected components. Conversely, the deletion of a node inherently triggers a cascading deletion of all its incident edges ($\text{Del}_V \to \text{Del}_E$) to avoid invalid dangling bonds. We further extend this to motif-level perturbations, such as replacing or hallucinating entire functional groups, which require a synchronized sequence of multiple additions/deletions.

To ensure varying levels of structural degradation, each ground-truth molecule is injected with 1 to 6 such logical errors. Due to the coordinated nature of composite edits, a single logical sub-structure error may expand into multiple discrete edit instructions in the target sequence $\Delta$. This targeted corruption strategy forces the VLM to go beyond trivial one-to-one symbol corrections, teaching it to perform multi-step, chemically plausible topological restorations.

\subsection{Data Synthesis for Corner Cases}
\begin{figure*}[t]
    \centering
    \includegraphics[width=1.0\textwidth]{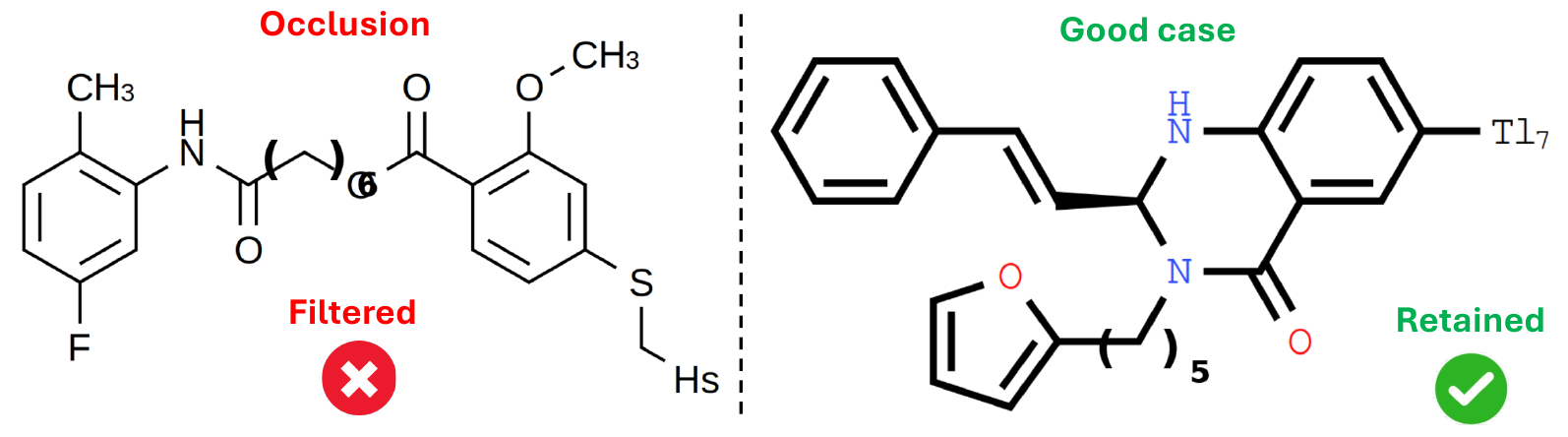}
    \vspace{-1.5em}
    \caption{Examples of human filtered and retained corner case training samples.}
    \label{fig:corner case suppl}
    \vspace{-1.5em}
\end{figure*}

\begin{figure*}[t]
    \centering
    \includegraphics[width=1.0\textwidth]{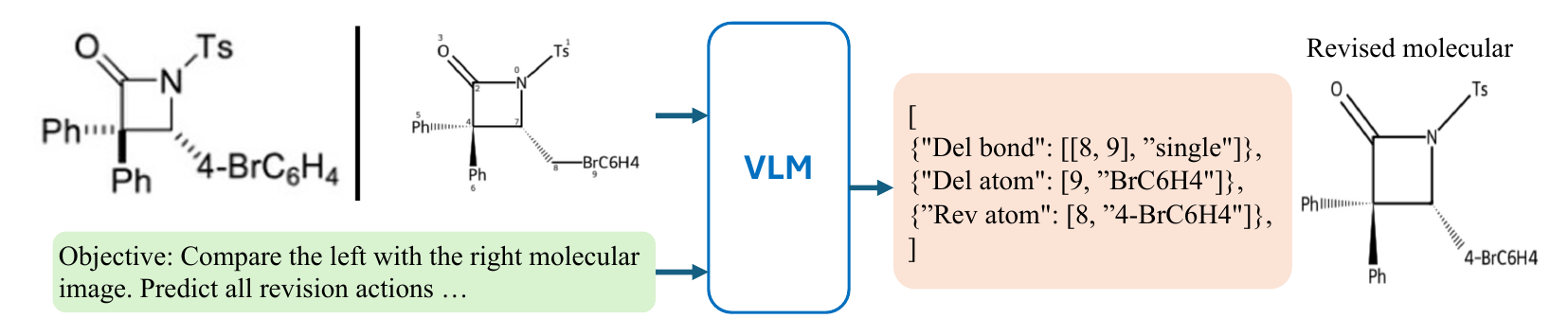}
    \vspace{-1.5em}
    \caption{An example of the input and output of the ViCoR for revision.}
    \label{fig:vlm example}
    \vspace{-1.5em}
\end{figure*}

Since Indigo does not support image synthesis for certain corner cases, we adopt a post-processing–based synthesis strategy combined with human filtering to construct such samples. As in Fig.~\ref{fig:corner case suppl}, we insert repeated brackets at the target atom locations using their coordinates in the synthesized images. The resulting samples are then manually filtered to remove visually implausible instances, yielding the final corner-case training data.
For corner cases involving special R-group symbols(Fig.~\ref{fig:examples}), we assign a dedicated label \texttt{r} to distinguish them from standard R-group symbols.

\section{Prompt for ViCoR.}
 We employ task-specific prompts for the ViCoR to support graph verification and revision, summarized in Tab.~\ref{tab:prompts}. It is worth noting that our VLM is task-specifically trained rather than direct prompt-engineering on an existing pre-trained VLM.

\begin{table}[!htbp]
\centering
\caption{Prompts used for verification and revision in ViCoR}
\label{tab:prompts}
\renewcommand{\arraystretch}{1.2}
\setlength{\tabcolsep}{6pt}
\begin{tabular}{p{0.18\linewidth} p{0.75\linewidth}}
\toprule
\textbf{Task} & \textbf{Prompt} \\
\midrule
Verification &
Do the left and right molecular images represent the same molecule? \newline
Ignore the drawing style and other noises. \newline
Answer only \texttt{`Yes.'} or \texttt{`No.'}. \\
\midrule
Revision &
Objective: \newline
Compare the left with the right molecular image. Predict all revision actions to reconstruct the exact graph shown in the reference image. \vspace{0.5em}\newline
Instructions: \newline
1. Check for missing, redundant, or incorrect atoms and bonds. \newline
2. Match indices to specific visual regions strictly using spatial proximity. \newline
3. Formulate the corrections as a sequence of discrete graph edit operations. \vspace{0.5em}\newline
Output Format: \newline
Return ONLY a JSON list of dictionaries containing the required edit instructions. You must use the following templates: \newline
\textbullet\ \texttt{\{"Add atom": [\textless index\textgreater, "\textless symbol\textgreater", ["\textless x\textgreater", "\textless y\textgreater"]]\}} \newline
\textbullet\ \texttt{\{"Del atom": [\textless index\textgreater, "\textless symbol\textgreater"]\}} \newline
\textbullet\ \texttt{\{"Rev atom": [\textless index\textgreater, "\textless target\_symbol\textgreater"]\}} \newline
\textbullet\ \texttt{\{"Add bond": [[\textless src\textgreater, \textless tgt\textgreater], "\textless bond\_type\textgreater"]\}} \newline
\textbullet\ \texttt{\{"Del bond": [[\textless src\textgreater, \textless tgt\textgreater], "\textless bond\_type\textgreater"]\}} \newline
\textbullet\ \texttt{\{"Rev bond": [[\textless src\textgreater, \textless tgt\textgreater], "\textless target\_bond\_type\textgreater"]\}} \vspace{0.5em}\newline
Example: \newline
\texttt{[\{"Add atom": [3, "C", ["\textless 0.314\textgreater", "\textless 0.576\textgreater"]]\}, \{"Add bond": [[3, 4], "solid wedge"]\}]} \\
\bottomrule
\end{tabular}
\vspace{-1em}
\end{table}

\section{Details on Index-anchored Executable Revision}
\label{sec:revision_details}

As introduced in the main text, the Vision-Language Model (VLM) predicts a sequence of graph edit operations $\Delta = (\delta_1, \delta_2, \dots, \delta_K)$ in a single forward pass. Because these instructions are generated autoregressively based on the initial input state, applying them simultaneously to the parsed graph can lead to logical collisions and topological invalidity. To manage overlapping edits and ensure structural robustness, the graph update operator $\oplus$ processes the sequence strictly in order, computing intermediate graph states $G_k = G_{k-1} \oplus \delta_k$ (where $G_0$ is the initial parsed graph). 

During this sequential execution, the $\oplus$ operator employs a dynamic validity enforcement mechanism to resolve conflicts through two core strategies:

\textbf{Absorbing Implicit Topological Changes.} 
Certain graph edits inherently trigger cascading effects that alter the topology beyond the explicit instruction. The $\oplus$ operator automatically absorbs these implicit changes to maintain chemical validity. The most prominent case is \textbf{cascading deletion}: when a node deletion instruction $\text{Del}_V(i)$ is executed, the operator not only removes atom $i$, but also automatically identifies and removes all incident edges $E(i, j)$ connected to atom $i$. This prevents the emergence of invalid ``dangling bonds'' (edges without endpoints) in the intermediate graph $G_k$.

\textbf{Bypassing Obsolete Instructions.} 
Because the target graph continuously evolves during the sequential update, some subsequent instructions $\delta_k$ may become invalid or obsolete relative to the current state $G_{k-1}$. Instead of raising execution errors, the $\oplus$ operator evaluates the pre-conditions of each $\delta_k$ and safely bypasses (ignores) it if conflicts are detected. Typical scenarios include:
\begin{itemize}
    \item \textbf{Target Missing (Invalid Edge Operations):} If an instruction attempts to add, delete, or revise an edge between nodes $i$ and $j$ (e.g., $\text{Add}_E(i \to j)$), but either node $i$ or $j$ has already been removed by a preceding $\text{Del}_V$ operation, the operator bypasses this edge instruction.
    \item \textbf{Redundant Operations:} If an instruction requests to delete an atom/bond that has already been implicitly deleted (via cascading deletion), or requests to revise a target to its current state, the operation is skipped.
    \item \textbf{Index Collisions:} If a node addition $\text{Add}_V(i)$ attempts to use an index $i$ that already exists in $G_{k-1}$, the operator dynamically assigns the next available unique index and updates subsequent edges pointing to this new node.
\end{itemize}

By coupling sequential execution with these conflict resolution protocols, the $\oplus$ operator guarantees that the final reconstructed graph $G_K$ is strictly valid in terms of graph topology, fully decoding the VLM's predictions into chemically sound molecular structures.

\section{Test Datasets}
To compare with prior state-of-the-art methods, we evaluate molecular structure recognition on six public realistic benchmarks: CLEF~\citep{piroi2010clef}, UOB~\citep{sadawi2012chemical}, JPO~\citep{fujiyoshi2011robust}, USPTO~\citep{filippov2009optical}, Staker~\citep{staker2019molecular}, and ACS~\citep{qian2023molscribe}, as well as two synthetic datasets (Indigo and ChemDraw~\citep{qian2023molscribe}). However, these classical OCSR benchmarks contain limited coverage of R-group symbols, complex abbreviations, and dissociative noise, which are frequently encountered in real-world chemical literature.

To better assess ViCoR under practical conditions, we curate two in-the-wild datasets, OCIE and JIE. OCIE consists of 1,141 molecular images cropped from the chemical information extraction test set of OpenChemIE~\citep{fan2024openchemie}. JIE contains 1,779 molecular images extracted from 243 figures collected from chemical literature sources such as \emph{JACS} and \emph{Nature}. All dataset statistics are summarized in Tab.~\ref{tab:test_datasets}.

\begin{table}[!htbp]
\centering
\footnotesize
\caption{
Comparison of OCIE and JIE with existing real-world OCSR benchmarks.
OCIE and JIE complement patent-domain benchmarks with recent journal figures containing smaller image crops, frequent color usage, and naturally co-occurring recognition challenges.
}
\label{tab}
\setlength{\tabcolsep}{4.5pt}
\renewcommand{\arraystretch}{1.08}
\resizebox{\linewidth}{!}{
\begin{tabular}{@{}lllrcc@{}}
\toprule
\textbf{Dataset}
& \textbf{Source domain}
& \textbf{Primary focus}
& \textbf{Images}
& \textbf{Median size}
& \textbf{Color (\%)} \\
\midrule

IP5-M
& Patents
& Markush / R-groups
& 878
& $1024\times1024$
& 0.2 \\

USPTO-10K-Abb
& Patents
& Abbreviated superatoms
& 10,000
& $648\times394$
& 0.0 \\

\midrule

OCIE
& Journal figures (2022--2023)
& Mixed OCSR factors
& 1,141
& $220\times155$
& 72.0 \\

JIE
& Journal figures (2022--2024)
& Mixed OCSR factors
& 1,779
& $233\times158$
& 19.6 \\

\bottomrule
\end{tabular}
}
\end{table}

\textbf{Complementarity to Existing Real-World Benchmarks.}
OCIE and JIE are designed to complement existing real-world OCSR benchmarks.
In particular, IP5-M focuses on Markush structures and R-groups in patent images, while USPTO-10K-Abb targets abbreviated superatoms in the patent domain.
OCIE and JIE instead provide an updated journal-domain evaluation with the following complementary properties:

\begin{itemize}
\item \textbf{Journal-domain visual distribution.}
Journal substrate-scope and reaction figures exhibit heterogeneous layouts, line widths, backgrounds, drawing conventions, and color usage that differ from predominantly monochrome patent drawings.

\item \textbf{Low-resolution crops with natural interference.}
OCIE and JIE consist of natural image crops with median resolutions of approximately $220\times155$ and $233\times158$ pixels, respectively.
Since individual molecules are cropped from complete journal figures, the images may retain neighboring structure fragments, labels, or other residual figure content, introducing realistic visual distractors.

\item \textbf{Naturally co-occurring recognition challenges.}
Rather than isolating a single difficulty, OCIE and JIE preserve the combinations naturally found in journal figures.
R-groups, abbreviations, stereochemistry, formal charges, and disconnected structures may co-occur with color, low resolution, heterogeneous drawing styles, and crop interference.

\end{itemize}

Together, these properties provide a complementary test setting for evaluating OCSR systems under realistic and visually heterogeneous journal conditions.

\textbf{Dataset Construction.} 
To construct the OCIE and JIE datasets, we compile a diverse corpus of scientific articles in PDF format from prominent chemistry and multidisciplinary journals (e.g., \emph{JACS}, \emph{Nature}). Figures are systematically extracted from these PDFs using VisualHeist, followed by the manual selection of reaction scope diagrams. We subsequently apply MolDetect~\citep{fan2024openchemie} to localize individual molecular bounding boxes. These boxes are then cropped to generate the single-molecule images that comprise our final datasets, ensuring a comprehensive representation of realistic molecular depictions.
For ground-truth annotation, we employ a semi-automatic expert curation pipeline. Initially, MolScribe is utilized to generate a candidate molecular graph for each image. These preliminary graphs are then rigorously reviewed and manually corrected by three independent annotators with advanced chemistry backgrounds. To guarantee high-quality and consistent annotations, each sample undergoes independent verification, and any inter-annotator disagreements are resolved through consensus discussions.

\begin{table}[!htbp]
\centering
\caption{Summary of the test datasets}
\label{tab:test_datasets}
\renewcommand{\arraystretch}{1.2}
\setlength{\tabcolsep}{6pt}
\begin{tabular}{l l r c}
\toprule
\textbf{Dataset} & \textbf{Type} & \textbf{Total Images} & \textbf{Abbreviations} \\
\midrule
Indigo~\citep{qian2023molscribe}   & Synthetic & 5,719  & $\times$ \\
ChemDraw~\citep{qian2023molscribe}& Synthetic & 5,719  & $\times$ \\
\midrule
CLEF\citep{piroi2010clef}    & Real      & 992    & $\checkmark$ \\
UOB~\citep{sadawi2012chemical}     & Real      & 5,740  & $\checkmark$ \\
JPO~\citep{fujiyoshi2011robust}     & Real      & 450    & $\times$ \\
USPTO~\citep{filippov2009optical}   & Real      & 5,719  & $\checkmark$ \\
Staker~\citep{staker2019molecular}  & Real      & 50,000 & $\checkmark$ \\
ACS~\citep{qian2023molscribe}     & Real      & 331    & $\checkmark$ \\
\midrule
OCIE    & Real      & 1141    & $\checkmark$ \\
JIE     & Real      & 1779    & $\checkmark$ \\
\bottomrule
\end{tabular}
\end{table}

\section{More Experiments and Ablation Study}
\label{app:more_ablation}

We provide additional analyses of the computational cost and design choices of
ViCoR that complement the ablations in the main text.
We first quantify the accuracy--efficiency trade-off introduced by iterative
verification and revision, and then study robustness to rendering styles and
the effect of the VLM backbone and model scale.

\textbf{Computational Cost and Iteration Budget.}
\label{app:efficiency}
We first compare the end-to-end computational cost of ViCoR with representative
OCSR systems under the same local inference environment.
All measurements are obtained on a single NVIDIA RTX 3090 and include the
complete inference pipeline.
For ViCoR, this includes the base recognizer, prediction rendering, VLM
verification and revision, and graph-update operations.
All reported accuracy differences are absolute percentage-point changes.

\begin{table*}[!htbp]
\centering
\footnotesize
\caption{
End-to-end accuracy and computational cost on a single NVIDIA RTX 3090.
Relative latency is normalized by MolScribe.
}
\label{tab:runtime_comparison}
\setlength{\tabcolsep}{4.5pt}
\renewcommand{\arraystretch}{1.08}
\resizebox{\textwidth}{!}{
\begin{tabular}{@{}lccccccc@{}}
\toprule
&
&
&
&
\multicolumn{2}{c}{\textbf{OCIE}}
&
\multicolumn{2}{c}{\textbf{JIE}} \\
\cmidrule(lr){5-6}
\cmidrule(lr){7-8}

\textbf{Method}
& \textbf{Throughput $\uparrow$}
& \textbf{Latency $\downarrow$}
& \textbf{Rel. latency}
& \textbf{OA $\uparrow$}
& \textbf{$\Delta$}
& \textbf{OA $\uparrow$}
& \textbf{$\Delta$} \\
&
\textbf{(img/s)}
& \textbf{(s/img)}
& \textbf{vs.\ MolScribe}
&
&
\textbf{vs.\ MolScribe}
&
&
\textbf{vs.\ MolScribe} \\
\midrule

OSRA
& 6.64
& 0.151
& 0.24$\times$
& 36.17
& $-37.36$
& 19.88
& $-41.95$ \\

DECIMER
& 0.28
& 3.571
& 5.75$\times$
& 35.07
& $-38.46$
& 16.42
& $-45.41$ \\

MolGrapher
& 0.42
& 2.381
& 3.83$\times$
& 41.20
& $-32.33$
& 25.40
& $-36.43$ \\

MolNexTR
& 1.42
& 0.704
& 1.13$\times$
& 73.24
& $-0.29$
& 61.62
& $-0.21$ \\

MolScribe
& 1.61
& 0.621
& 1.00$\times$
& 73.53
& --
& 61.83
& -- \\

\textbf{MolScribe + ViCoR}
& 1.18
& 0.847
& 1.36$\times$
& \textbf{88.26}
& \textbf{+14.73}
& \textbf{84.32}
& \textbf{+22.49} \\

\bottomrule
\end{tabular}
}
\end{table*}

As shown in Tab.~\ref{tab:runtime_comparison}, ViCoR introduces a moderate
latency increase over MolScribe while providing substantially higher strict
recognition accuracy.
The default $T=3$ configuration increases latency by $1.36\times$ relative to
MolScribe, but improves OA by 14.73 and 22.49 points on OCIE and JIE,
respectively.
ViCoR is also considerably faster than DECIMER and MolGrapher while achieving
much higher accuracy.
OSRA provides the highest throughput, but at a substantial loss in recognition
quality.
These results indicate that ViCoR trades a moderate amount of additional
computation for a large improvement in reliable whole-structure recognition.

We further isolate the additional cost introduced by iterative verification
and revision by varying the maximum number of iterations $T$.

\begin{table*}[!htbp]
\centering
\footnotesize
\caption{
Accuracy--cost trade-off under different ViCoR iteration budgets.
Relative latency is normalized by the MolScribe base recognizer.
}
\label{tab:iteration_runtime}
\setlength{\tabcolsep}{5pt}
\renewcommand{\arraystretch}{1.08}
\resizebox{0.92\textwidth}{!}{
\begin{tabular}{@{}lccccccc@{}}
\toprule
&
&
&
&
\multicolumn{2}{c}{\textbf{OCIE}}
&
\multicolumn{2}{c}{\textbf{JIE}} \\
\cmidrule(lr){5-6}
\cmidrule(lr){7-8}

\textbf{Method}
& \textbf{Throughput $\uparrow$}
& \textbf{Latency $\downarrow$}
& \textbf{Rel. latency}
& \textbf{OA $\uparrow$}
& \textbf{$\Delta$}
& \textbf{OA $\uparrow$}
& \textbf{$\Delta$} \\
&
\textbf{(img/s)}
& \textbf{(s/img)}
& \textbf{vs.\ base}
&
&
\textbf{vs.\ base}
&
&
\textbf{vs.\ base} \\
\midrule

MolScribe base
& 1.61
& 0.621
& 1.00$\times$
& 73.53
& --
& 61.83
& -- \\

ViCoR ($T=1$)
& 1.33
& 0.752
& 1.21$\times$
& 80.82
& +7.29
& 72.45
& +10.62 \\

ViCoR ($T=2$)
& 1.21
& 0.826
& 1.33$\times$
& 86.45
& +12.92
& 81.36
& +19.53 \\

\textbf{ViCoR ($T=3$)}
& 1.18
& 0.847
& 1.36$\times$
& \textbf{88.26}
& \textbf{+14.73}
& \textbf{84.32}
& \textbf{+22.49} \\

\bottomrule
\end{tabular}
}
\end{table*}

Tab.~\ref{tab:iteration_runtime} shows a clear accuracy--cost trade-off.
Increasing the iteration budget consistently improves OA, while the
incremental latency becomes progressively smaller.
Notably, $T=1$ already recovers substantial accuracy at only $1.21\times$ the
base latency, whereas the default $T=3$ setting provides the strongest
recognition performance at $1.36\times$ latency.
The iteration budget can therefore be selected according to the deployment
requirement: fewer iterations favor high-throughput extraction, while additional
iterations are beneficial when strict structural correctness is prioritized.

\textbf{Robustness to Rendering Style.}
\label{app:renderer_ablation}
ViCoR compares a real molecular image against a rendered prediction.
A natural concern is therefore whether the verifier learns renderer-specific
appearance rather than structural correspondence.
We test this by comparing training pairs generated with the same renderer
(RDKit $\rightarrow$ RDKit) against cross-renderer training
(Indigo $\rightarrow$ RDKit).

\begin{table}[!htbp]
\centering
\small
\caption{
Effect of rendering diversity during verification training (\%).
}
\label{tab:renderer_ablation}
\setlength{\tabcolsep}{7pt}
\renewcommand{\arraystretch}{1.08}
\begin{tabular}{@{}lcc@{}}
\toprule
\textbf{Training comparison}
& \textbf{OCIE OA $\uparrow$}
& \textbf{JIE OA $\uparrow$} \\
\midrule

RDKit $\rightarrow$ RDKit
& 85.42
& 80.76 \\

Indigo $\rightarrow$ RDKit
& \textbf{88.26}
& \textbf{84.32} \\

\bottomrule
\end{tabular}
\end{table}

Cross-renderer training improves OA by 2.84 and 3.56 points on OCIE and JIE,
respectively.
This suggests that exposing the verifier to depiction shifts during training
reduces reliance on renderer-specific visual cues and encourages comparison
based on the underlying molecular structure.
We therefore use cross-renderer training in the default ViCoR configuration.

\textbf{Effect of VLM Backbone and Scale.}
\label{app:vlm_scale}
We next examine whether the performance of ViCoR primarily arises from using a
larger general-purpose VLM or from task-specific verification--revision
training.
We compare two general-purpose GPT models with trained Qwen2.5-VL models at
3B and 7B parameter scales.

\begin{table}[!htbp]
\centering
\small
\caption{
Effect of the VLM backbone and model scale on strict recognition accuracy
(\%).
}
\label{tab:vlm_scale}
\setlength{\tabcolsep}{7pt}
\renewcommand{\arraystretch}{1.08}
\begin{tabular}{@{}lcc@{}}
\toprule
\textbf{VLM}
& \textbf{OCIE OA $\uparrow$}
& \textbf{JIE OA $\uparrow$} \\
\midrule

ViCoR(GPT-4o) + MolScribe
& 76.60
& 73.97 \\

ViCoR(GPT-5.6-sol) + MolScribe
& 77.83
& 75.77 \\

ViCoR(Trained Qwen2.5-VL-3B) + MolScribe
& 88.26
& 84.32 \\

ViCoR(Trained Qwen2.5-VL-7B) + MolScribe
& \textbf{88.48}
& \textbf{84.53} \\

\bottomrule
\end{tabular}
\end{table}

As shown in Tab.~\ref{tab:vlm_scale}, increasing the trained Qwen2.5-VL
backbone from 3B to 7B parameters yields only marginal gains of 0.22 and
0.21 points on OCIE and JIE.
In contrast, both trained models substantially outperform the general-purpose
GPT baselines.
This indicates that, within the tested scale range, ViCoR benefits more from
task-specific verification--revision training than from simply increasing VLM
parameter count.
We therefore use Qwen2.5-VL-3B as the default verifier/reviser to balance
accuracy and computational efficiency.

\section{Visualization.}
\begin{figure*}[!htbp]
    \centering
    \includegraphics[width=1.0\textwidth]{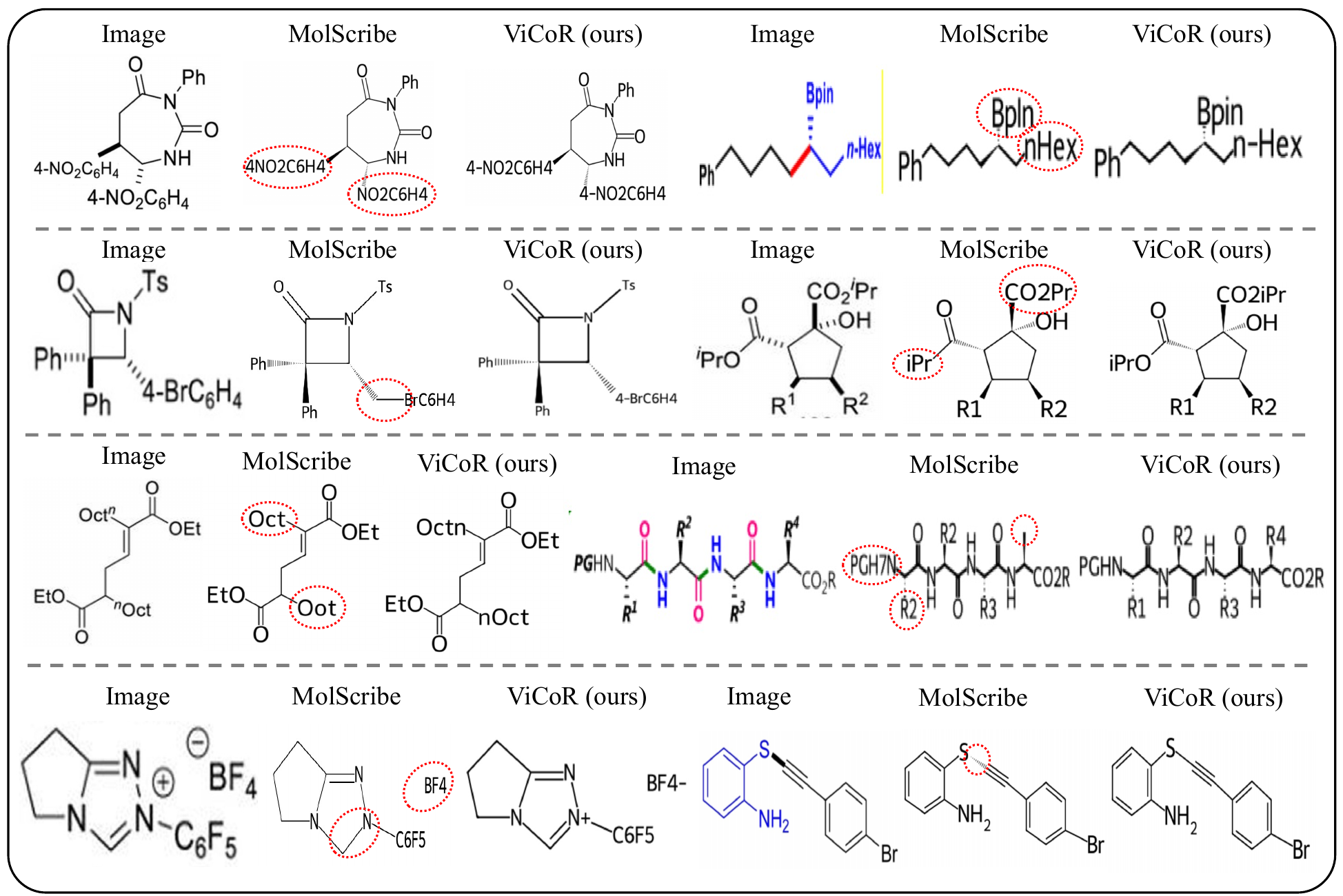}
    \caption{Examples of comparison between baseline MolScribe and MolScribe + ViCoR.}
    \label{fig:more examples}
\end{figure*}

Fig.~\ref{fig:more examples} illustrates more examples of molecular structure recognition results by MolScribe and MolScribe + ViCoR.

\section{Downstream 1: Chemical Reaction Extraction}
\begin{figure*}[!htbp]
    \centering
    \includegraphics[width=1.0\textwidth]{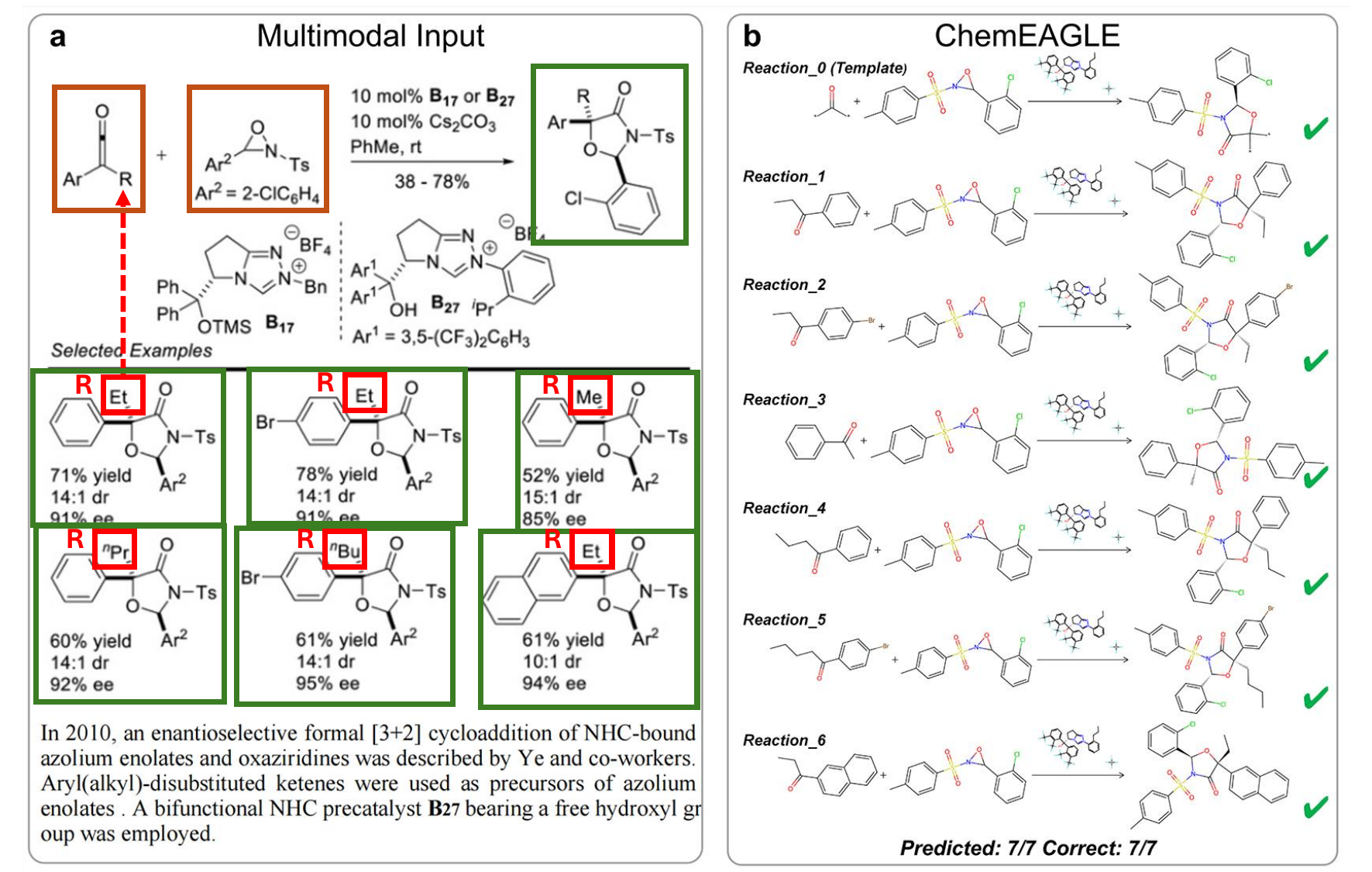}
    \caption{Example of chemical information extraction (Left: input reaction scope image. Right: expected extracted 7 reactions from the image). This figure comes from~\citep{chen2025multi}.}
    \label{fig:chem info extract}
\end{figure*}

An important application of ViCoR is chemical reaction extraction from scientific literature. Chemical reaction extraction aims to extract structured reactions from complex reaction scope figures, which typically requires a multi-stage pipeline including \textbf{reaction template recognition}, \textbf{recognition of multiple representative product molecules}, and \textbf{R-group substitution between products and reaction templates}. This process is highly error-sensitive, as failures at any stage can propagate and result in invalid or incorrect reactions, degrading the performance of the downstream tasks trained on the extracted reactions.
An example of chemical reaction extraction is shown in Fig.~\ref{fig:chem info extract}.

\section{Downstream 2: Reaction Prediction with Literature-Extracted Data}
\label{app:reaction_prediction}

We further evaluate whether improvements in selective structure recognition translate into better downstream model learning by constructing reaction-prediction datasets directly from figures in the chemical literature.
All molecular and reaction images in this experiment are cropped from original journal articles.

\textbf{Literature Data Collection.}
We collect reaction figures from papers published in \textit{The Journal of Organic Chemistry}, \textit{Organic Letters}, and \textit{Journal of the American Chemical Society (JACS)} between 2000 and 2025.
The resulting corpus contains 2,100 figures related to photocatalytic reactions and 430 figures related to Suzuki coupling.
To avoid leakage across visually or chemically related figures from the same publication, we split the corpus into training and test sets at the article level using an 80/20 split.

\textbf{Reaction Extraction and SSR Processing.}
We apply OpenChemIE~\citep{fan2024openchemie} to each literature figure to recover structured reaction records.
OpenChemIE first detects molecules and reaction arrows and establishes their spatial relationships, after which its OCSR module converts each localized molecular image into a molecular structure.
The recognized structures are then assembled according to the detected reaction layout to obtain complete reactant--product pairs.
Using raw OpenChemIE extraction, the training split contains 25,659 photocatalytic reactions and 4,233 Suzuki-coupling reactions.

\textbf{Expert-Corrected Evaluation Sets.}
To ensure that downstream evaluation is independent of OCSR errors, we construct fixed reference test sets from the held-out literature figures.
Three domain experts manually inspect and correct the automatically extracted reactions, including molecular structures, reactant--product assignments, and reaction boundaries when necessary.
After correction, the test sets contain 4,291 photocatalytic reactions and 668 Suzuki-coupling reactions.
The same expert-corrected test sets are used to evaluate models trained under all extraction strategies.

\textbf{Downstream Reaction Prediction.}
For each training corpus, we fine-tune the same Qwen3-8B model~\citep{yang2025qwen3} for forward reaction prediction, where the input consists of the extracted reactants and the target is the corresponding product.
All model architecture, optimization settings, and downstream evaluation procedures are kept identical across conditions; the only difference is the post-recognition strategy used to construct the training reactions.
We evaluate prediction quality using exact-match accuracy together with molecular similarity metrics based on RDKit and Morgan fingerprints.
This experiment isolates the effect of extraction quality on downstream learning.
Because a single OCSR error can alter molecular connectivity, atom identity, or bond order, erroneous structures may convert an otherwise valid literature reaction into an incorrect training pair.
The results in Tab.~\ref{tab:reaction_prediction} show that improving both the reliability and coverage of extracted structures leads to consistently better reaction-prediction models.

\section{Transfer to BPMN Process-Graph Recognition}
\label{app:bpmn}

To examine whether ViCoR is specific to molecular diagrams, we transfer its
index-anchored render--verify--revise interface to hand-drawn BPMN
process-graph recognition.
The key idea remains unchanged: a base parser first predicts a structured graph,
which is rendered into a spatially aligned and explicitly indexed visual
reference; the VLM then verifies the prediction and produces localized,
executable node/edge edits rather than regenerating the entire graph.

\textbf{Dataset and task.}
We use hdBPMN v1.0.0 from Sketch2Process
~\citep{schafer2022sketch2process} and follow its official writer-disjoint split,
which contains 704 images in total.
We consider structure-only image-to-graph recognition with eight BPMN node
classes and directed sequence-flow edges, excluding text, OCR, and other
document metadata.
Faster R-CNN~\citep{ren2016faster} and YOLO11m~\citep{khanam2024yolov11}
serve as two representative base parsers.

\textbf{Training and evaluation protocol.}
For each base parser, correction candidates for training are generated through
three-fold writer-disjoint out-of-fold inference, preventing the verifier from
seeing predictions produced on images used to train the corresponding parser.
Decision thresholds are selected exclusively on the validation split, and the
test set is evaluated once.
We report the same reliability-oriented metrics used in the main experiments.
In particular, OA requires strict whole-graph exact match after Hungarian
node matching, such that all evaluated node labels and directed edges must be
correct simultaneously.

\begin{table}[t]
\centering
\small
\caption{
Transfer of ViCoR to hand-drawn BPMN process-graph recognition (\%).
Results are reported on the official writer-disjoint split of hdBPMN v1.0.0.
}
\label{tab:bpmn_transfer}
\setlength{\tabcolsep}{6pt}
\renewcommand{\arraystretch}{1.08}
\begin{tabular}{@{}lccc@{}}
\toprule
\textbf{Method}
& \textbf{OA $\uparrow$}
& \textbf{AA $\uparrow$}
& \textbf{Cov $\uparrow$} \\
\midrule
Faster R-CNN base
& 41.60 & 41.60 & 100.00 \\
YOLO11m base
& 50.40 & 50.40 & 100.00 \\
\midrule
Faster R-CNN + ViCoR
& 62.10 & 95.50 & 57.80 \\
YOLO11m + ViCoR
& 68.80 & 96.30 & 64.80 \\
\bottomrule
\end{tabular}
\end{table}

\textbf{Results.}
As shown in Tab.~\ref{tab:bpmn_transfer}, ViCoR substantially improves strict
whole-graph recognition with both base parsers.
OA increases from 41.6\% to 62.1\% with Faster R-CNN and from 50.4\% to
68.8\% with YOLO11m, corresponding to gains of 20.5 and 18.4 percentage points,
respectively.
At the same time, ViCoR achieves high accepted accuracy of 95.5\% and 96.3\%,
while retaining 57.8\% and 64.8\% coverage.
The consistent improvements across two different parsers indicate that the
verification--revision interface is not tied to a particular detection
architecture.
Since OA requires the entire predicted graph to match exactly, these gains
reflect graph-level structural correction rather than improvements to isolated
node detections.
Overall, the results suggest that spatially aligned verification and
index-anchored revision can transfer beyond OCSR to other image-to-structure
tasks with explicit node--edge structure.

\section{Model Details}

ViCoR adopts Qwen2.5-VL-3B~\citep{bai2025qwen2} as the backbone VLM. We also summarize the architecture used in MolScribe. The detailed configurations of image encoders and language/decoder modules are reported in Tab.~\ref{tab:model_details}.

\begin{table}[!htbp]\footnotesize
\centering
\setlength{\tabcolsep}{6pt}
\renewcommand{\arraystretch}{1.2}
\caption{Model architecture details.}
\label{tab:model_details}
\begin{tabular}{lcc}
\toprule
 & \textbf{Qwen2.5-VL-3B} & \textbf{MolScribe} \\
\midrule
\multicolumn{3}{l}{\emph{Image Encoder}} \\
Architecture & Vision Transformer & Swin Transformer (Swin-B) \\
Layers & 32 & 36 \\
Hidden Size & 1280 & -- \\
Attention Heads & 16 & -- \\
Patch Size & 14 & -- \\
Parameters & -- & 88M \\
Pre-training & -- & ImageNet-22K \\
Input Resolution & -- & $384 \times 384$ \\
\midrule
\multicolumn{3}{l}{\emph{Language / Decoder Module}} \\
Architecture & Transformer LLM & Transformer Decoder \\
Layers & 36 & 6 \\
Hidden Size & 2048 & 256 \\
Attention / KV Heads & 2 (KV heads) & 8 \\
Head Size & 128 & -- \\
Positional Encoding & Rotary (LLM) & Sinusoidal \\
Dropout & -- & 0.1 \\
\midrule
\multicolumn{3}{l}{\emph{Prediction Head}} \\
Bond Predictor & -- & 2-layer FFN (ReLU) \\
\bottomrule
\end{tabular}
\end{table}

\end{document}